\documentclass[review]{elsarticle}
\usepackage{caption}
\usepackage{color}
\usepackage{multirow}
\usepackage{lineno,hyperref}
\usepackage{array}
\usepackage{amsmath}
\usepackage{booktabs}
\usepackage[subrefformat=parens,labelformat=parens]{subfig}
\usepackage{amssymb}
\usepackage[all]{nowidow}
\usepackage{longtable}
\usepackage[title]{appendix}

\modulolinenumbers[5]

\journal{Pattern Recognition}

\biboptions{numbers,sort&compress}
\begin{document}

\begin{frontmatter}
\title{Deep Attentive Time Warping}

\author[kyushu_univ]{Shinnosuke~Matsuo\corref{mycorrespondingauthor}}
\cortext[mycorrespondingauthor]{Corresponding author}
\ead{shinnosuke.matsuo@human.ait.kyushu-u.ac.jp}

\author[ntt]{Xiaomeng~Wu}
\author[helsinki_univ]{Gantugs~Atarsaikhan}
\author[ntt]{Akisato~Kimura}
\author[ntt]{Kunio~Kashino}
\author[kyushu_univ]{Brian~Kenji~Iwana}

\author[kyushu_univ]{Seiichi~Uchida}
\ead{uchida@ait.kyushu-u.ac.jp}

\address[kyushu_univ]{Department of Advanced Information Technology, Kyushu University, Fukuoka, Japan}
\address[ntt]{Communication Science Laboratories, NTT Corporation, Kanagawa, Japan}
\address[helsinki_univ]{Institute for Molecular Medicine Finland, HiLIFE, University of Helsinki, Helsinki, Finland}

\begin{abstract}
Similarity measures for time series are important problems for time series classification. To handle the nonlinear time distortions, Dynamic Time Warping (DTW) has been widely used. However, DTW is not learnable and suffers from a trade-off between robustness against time distortion and discriminative power. In this paper, we propose a neural network model for task-adaptive time warping. Specifically, we use the attention model, called the bipartite attention model, to develop an explicit time warping mechanism with greater distortion invariance. Unlike other learnable models using DTW for warping, our model predicts all local correspondences between two time series and is trained based on metric learning, which enables it to learn the optimal data-dependent warping for the target task. We also propose to induce pre-training of our model by DTW to improve the discriminative power. Extensive experiments demonstrate the superior effectiveness of our model over DTW and its state-of-the-art performance in online signature verification.
\end{abstract}

\end{frontmatter}


\section{Introduction\label{sec:intro}}
Measuring similarity is one of the most important tasks for time series recognition. For example, similarity gives an essential criterion for classifying time series. Many applications, such as activity recognition, computational auditory scene analysis, computer security, electronic health records, and biometrics (e.g., handwritten signature verification)~\cite{FawazFWIM19}, use time series similarity for recognition. One difficulty in measuring similarity is due to nonlinear time distortions. The distortions can appear as temporal shifts, stretches and contractions, and other various nonlinear temporal fluctuations.\par
\begin{figure}[b]
\centering
\includegraphics[width=1.\columnwidth]{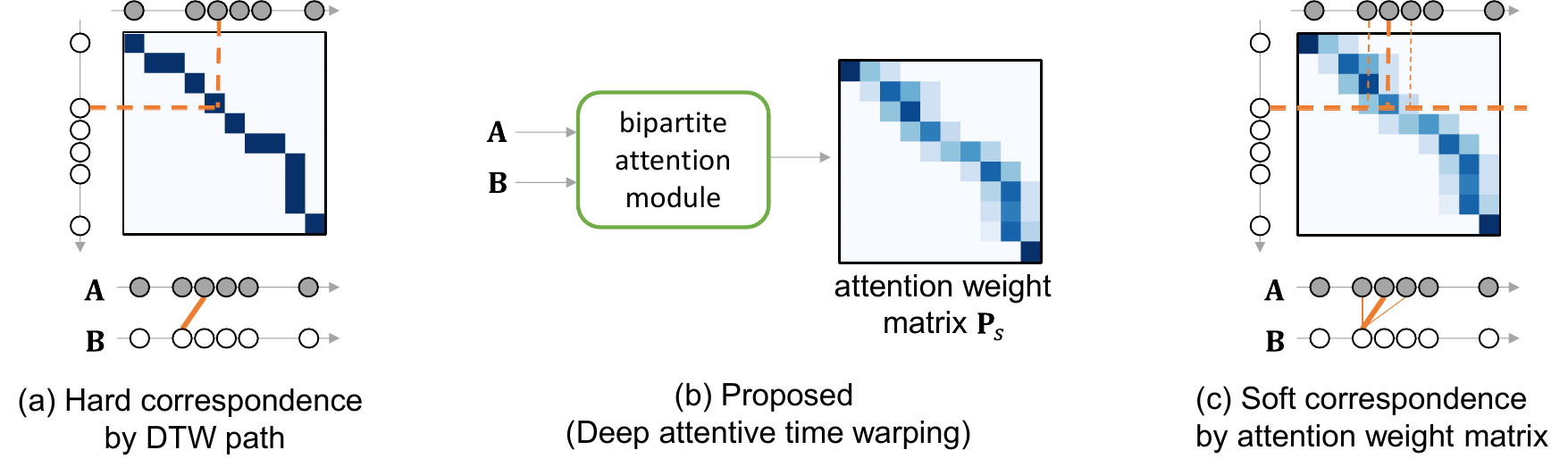}
\vspace{-10mm}
\caption{(a)~DTW conducts a hard correspondence between two sequences $\mathbf{A}$ and $\mathbf{B}$. (b)~The proposed {\em deep attentive time warping} is composed of a bipartite attention module, which generates an attention weight matrix $\mathbf{P}_s$. (c)~The attention weight matrix $\mathbf{P}_s$ represents soft correspondence between two sequences $\mathbf{A}$ and $\mathbf{B}$.}
\label{fig:overview}
\end{figure}
To be invariant to nonlinear time distortions, Dynamic Time Warping (DTW)~\cite{DTW78} has been widely utilized. Let $\mathbf{A}=\mathbf{a}_1,\ldots, \mathbf{a}_i,\ldots, \mathbf{a}_I$ and $\mathbf{B}=\mathbf{b}_1,\ldots, \mathbf{b}_j,\ldots, \mathbf{b}_J$ denote two time series, where both $\mathbf{a}_i$ and $\mathbf{b}_j$
are $D$-dimensional feature vectors.
As shown in Fig.~\ref{fig:overview}(a), DTW establishes a ``hard'' correspondence between $\mathbf{A}$ and $\mathbf{B}$ as a path on a two-dimensional plane, or an $I\times J$ binary matrix. Here the term ``hard'' implies ``one or zero;'' the $(i,j)$-th element of the matrix becomes 1, if $\mathbf{a}_i$ and $\mathbf{b}_j$ is ``matched (i.e., corresponding),'' and zero, otherwise. The correspondence is determined to minimize the distance  $\mathbf{A}$ and $\mathbf{B}$ by dynamic programming (DP).\par
In DTW, several hand-crafted constraints are often assumed for controlling the correspondence. Traditionally, monotonicity, continuity, and boundary constraints have been utilized. These constraints are appropriate and acceptable in many applications but encounter trade-off problems in the warping flexibility. If the constraints are too loose, DTW causes ``over-warping'' that cancels important inter-class differences and loses its discriminative power. If we add more constraints heuristically (like \cite{DTW_ww3,DTW_ww4}) to avoid over-warping, DTW then can not remove intra-class distortions sufficiently. 
\par

In recent years, {\em deep metric learning} has been applied to various classification tasks. Deep metric learning is a machine learning technique to learn an adaptive feature space that takes into account the similarity (or dissimilarity) relationships among data~\cite{Elad15,Chen17,Song17,Kaya2019DeepML}.
In the typical formulation, a Siamese or triplet neural network is trained to learn an embedding space, where closeness between embeddings (i.e., features extracted from the network) encodes the level of similarities between the data samples. It enforces the embeddings to lie close if the samples belong to the same class, and pushes them apart if different.\par
Deep metric learning for image classification tasks has had many successful results~\cite{Liu17, Wang18, Deng19}. 
Whereas for time series, there is still room for improvement.
As detailed in Section~\ref{sec:related}, the past attempts either suffer from the loss of useful temporal information~\cite{MuellerT16,CoskunTCNT18,RoyMM18} or are not explicitly invariant to nonlinear time distortions~\cite{BromleyGLSS93,TolosanaVFO18,AhrabianB19}. \par
In this paper, we propose a novel neural network-based time warping model, called {\em deep attentive time warping}\footnote{The code is available at {\tt https://github.com/matsuo-shinnosuke/deep-attentive-time\\ {\tt -warping}.}}. The proposed method is based on a novel learnable time warping mechanism with contrastive metric learning. Its key idea is a novel attention model, called the {\em bipartite attention module}.
As shown in Fig.~\ref{fig:overview}~(b), this module takes two time series as inputs and predicts an {\em attention weight matrix}. This matrix represents the ``soft'' correspondence between all time steps of the two inputs, as shown in  Fig.~\ref{fig:overview}~(c).
By training the bipartite attention module appropriately for a specific task, we can realize time warping that can mitigate the trade-off between robustness against time distortion and discriminative power. In other words, the learned soft correspondence will enhance important inter-class differences and, at the same time, remove intra-class distortions.\par
\begin{figure}[t]
\centering
\vspace{5mm}
\includegraphics[width=1.\columnwidth]{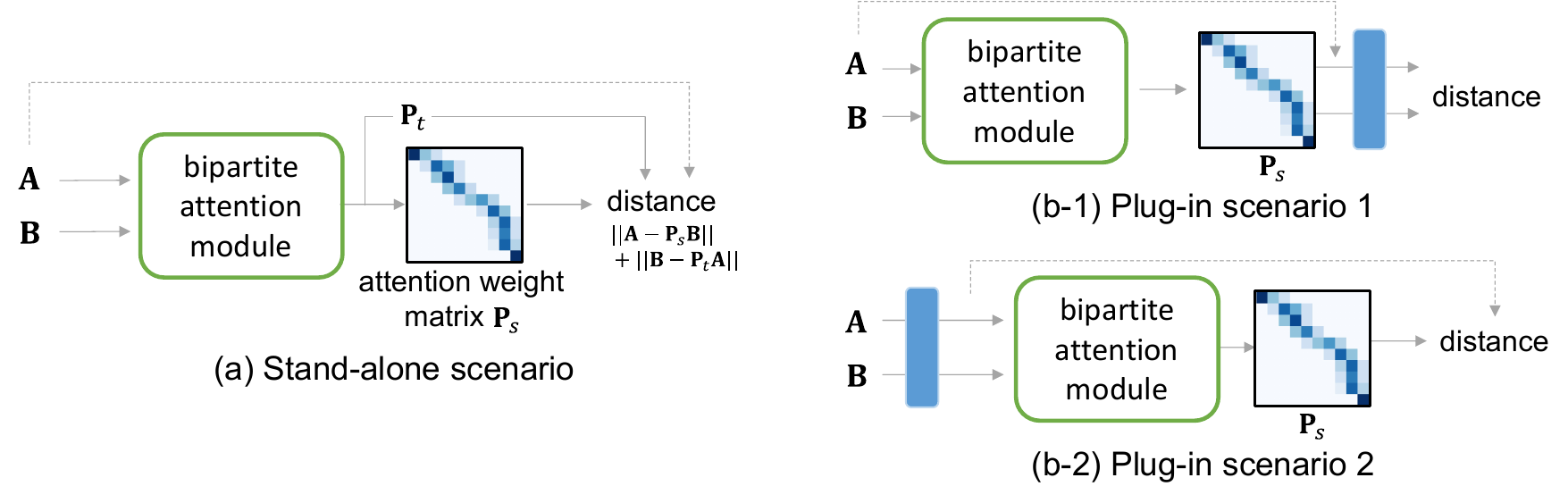}
\vspace{-10mm}
\caption{The proposed deep attentive time warping can be used in two scenarios. One is a stand-alone scenario~(a), and the other is a plug-in scenario~(b-1) and (b-2). In (b-1) and (b-2), blue boxes represent a neural network for contrastive representation learning.}
\label{fig:scenarios}
\end{figure}
The proposed method has great versatility and can be used for applications in two different scenarios; one is a stand-alone scenario and the other is a plug-in scenario. As shown in Fig.~\ref{fig:scenarios}~(a), the former takes two inputs  $\mathbf{A}$ and $\mathbf{B}$ and determines their difference by utilizing their original feature representation. In the latter scenario, we use existing contrastive metric learning frameworks with the standard DTW. Then, as shown in Figs.~\ref{fig:scenarios}~(b-1) and (b-2), we replace the DTW with the proposed method. Consequently, our deep attentive time warping is combined with contrastive representation learning and the entire framework becomes totally trainable for better (i.e., contrastive) time warping and feature representation. \par
We conduct extensive experiments to demonstrate the superior effectiveness of the proposed method.
We first conduct two experiments in the stand-alone scenario to confirm how the proposed method provides reasonable time warping for the classification. We prove that the proposed method achieves better classification performance with effective warping than the other time warping techniques through qualitative and quantitative evaluations on the well-known Unipen~\cite{Unipen} and the University of California Riverside (UCR)~\cite{UCRArchive} datasets. We then conduct another experiment in the plug-in scenario. Specifically, through an online signature verification experiment, we prove that the proposed method achieves state-of-the-art performance by outperforming other learnable time warping methods.\par

The main contributions of this paper are summarized as follows:
\begin{itemize}

\item A novel neural network-based time warping method, called deep attentive time warping, is proposed by introducing a bipartite attention module.
It is learnable, task-adaptive, and improves the trade-off between robustness against time distortion and discriminative power. We also prove a two-step training process enhances the performance.
\item We show the high versatility of the proposed deep attentive time warping by using it in two different scenarios, stand-alone and plug-in. 
\item Extensive experiments on more than 50 public datasets demonstrate the superior effectiveness of the proposed method over DTW as a stand-alone time warping model.
\item We experimentally show that the proposed method in the plug-in scenario achieves better performance than state-of-the-art learnable time warping methods in an online signature verification task.
\end{itemize}
%

\section{Related Work\label{sec:related}}

\subsection{Dynamic time warping\label{subsec:dtw}}

DTW~\cite{DTW78} (standard DTW) is a time warping method that has been used for a long time as a time series similarity measure invariant to nonlinear time distortion. As noted in Section~\ref{sec:intro}, DTW can determine the hard correspondence between $\mathbf{A}$ and $\mathbf{B}$. While DTW exhibits great distortion invariance, it may cause over-warping that often results in incorrect classification.\par

There are many attempts to improve the performance of DTW. To suppress over-warping, early studies~\cite{DTW78,DTW_ww3} proposed to put a warping window as an additional constraint to the standard monotonicity and continuity constraints. Roughly speaking, $\mathbf{a}_i$ is able to match with one of $\mathbf{b}_{i-w},\dots, \mathbf{b}_{i},\dots, \mathbf{b}_{i+w}$, where $w$ is the window width. A smaller $w$ will have fewer over-warping cases.
In~\cite{WDTW,Marteau09},  the warping path is penalized by the difference of $i$ and $j$ of the matched $\mathbf{a}_i$ and $\mathbf{b}_j$. Soft-DTW~\cite{soft-DTW} is an interesting attempt to replace the min operation with a soft-min operation, which is realized by logarithmic and exponential functions. With this replacement, DTW becomes differentiable and can be built in various machine learning frameworks. 

\subsection{DTW with deep metric learning\label{subsec:dtw_and_ml}}
In recent years, more efforts have been made on deep metric learning for time series~\cite{MuellerT16,CoskunTCNT18,RoyMM18,BromleyGLSS93,TolosanaVFO18,AhrabianB19}. They are based on a feature extraction mechanism with a Siamese network, which is trained by a loss function evaluating the distance between the features. The extracted features from time series are either global or local. Compared to the standard DTW, these methods achieve better accuracy; however, they do not treat the temporal distortion explicitly. This means that they do not warp one time series to another and, thus, is impossible to introduce an explicit control of warping flexibility.\par
%
Several metric learning methods introduce DTW for an explicit removal of temporal distortions. More specifically, they introduce the standard DTW before or after a Siamese network. Prewarping Siamese Network (PSN)~\cite{wu2019psn} and Time Aligned Recurrent Neural Networks (TARNN)~\cite{tarnn21} perform DTW between two time series and then fed the warped result to a Siamese network for metric learning. In contrast, Deep DTW (DDTW)~\cite{wu2019ddtw} first extracts a sequence of local features from each time series and then performs DTW. With the introduction of DTW, these methods could achieve better performance than simple metric learning methods. Note that they do not learn the warping characteristics; their temporal distortion removal ability relies on the standard DTW.\par

%
Few methods~\cite{CheHXL17,NeuralWarp} have been proposed for learning warping characteristics. They calculate a quasi-binary {\em matchability} $\Phi(i,j)$ between each $(i,j)$ pair. Then, all $IJ$ point-wise distances between $i\in [1,I]$ and $j\in [1,J]$ are aggregated by using $\Phi(i,j)$; if $\Phi(i,j)\sim 1$, the distance between $i$ and $j$ is taken into account. Since the matchability is determined {\em independently} for each $(i,j)$ pair by a neural network, 
it is time-consuming to have all $IJ$ matchability results. More importantly, this independent determination process cannot control the global warping characteristics, which have been carefully treated even in the standard DTW. 
\par
Furthermore, there is a preliminary conference paper of this work~\cite{matsuo2021attention} and this paper contains significant differences from it. First, we newly propose a plug-in scenario, where our deep attentive time warping is utilized as a differentiable module in a large classification system. We further confirmed that the plug-in usage of our technique achieves the state-of-the-art performance in large-scale signature verification tasks. Moreover, for the stand-alone scenario, we conduct more extensive comparison experiments on over 50 classification tasks in UCR dataset, whereas only four tasks have been tackled in~\cite{matsuo2021attention}. Technical details are also newly elaborated in this paper.
\par
\section{Deep Attentive Time Warping\label{sec:method}}
\subsection{Overview}


We propose {\em deep attentive time warping}, a novel neural network-based time warping method. As noted in Section~\ref{sec:intro}, the proposed method can be used to evaluate the distance/dissimilarity between two time series (e.g., series of raw signals or deep features) $\mathbf{A}$ and $\mathbf{B}$ in its stand-alone scenario of Fig.~\ref{fig:scenarios}~(a). It also can be used as an attention-based feature extractor in its plug-in scenario, as shown in Fig.~\ref{fig:scenarios}~(b).
\par
As shown in Figs.~\ref{fig:overview}~(b) and (c), the {\em bipartite attention module}  generates the attention weight matrix $\mathbf{P}_s$, which represents time warping between $\mathbf{A}$ and $\mathbf{B}$ as a soft temporal correspondence. The bipartite attention module is trained by metric learning with contrastive loss. The resulting matrix $\mathbf{P}_s$ is expected to provide not only distortion invariance but also discriminative power, both of which are appropriate for the target task.

\subsection{Time warping with the bipartite attention module\label{subsec:attention}}

\begin{figure}[t]
\centering
\includegraphics[width=.9\columnwidth]{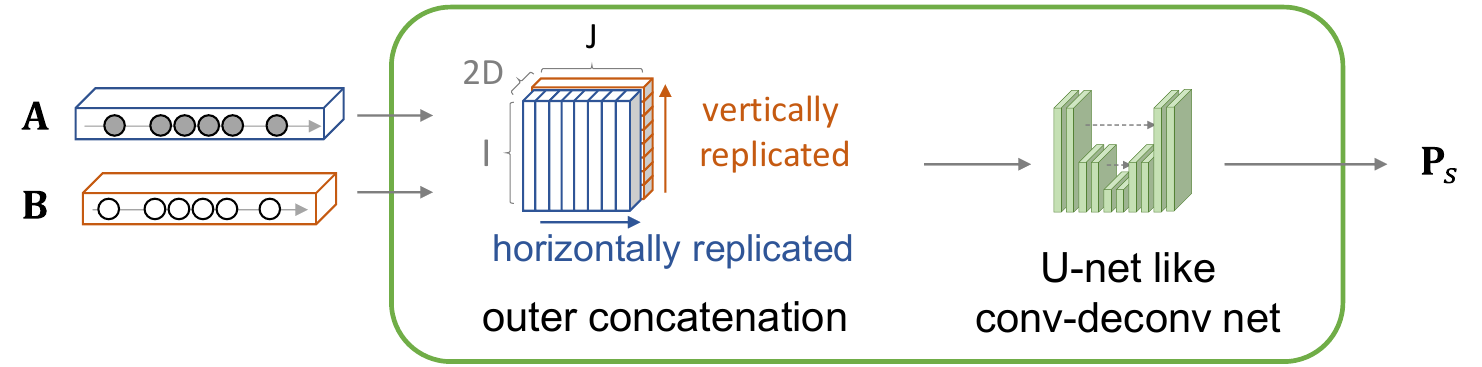}
\caption{Overview of the bipartite attention module.}
\label{fig:bipartite_attention_module}
\end{figure}

The detail of the bipartite attention module is shown in Fig.~\ref{fig:bipartite_attention_module}. In the bipartite attention module, 
two multivariate time series $\mathbf{A}$ and $\mathbf{B}$ are first combined by ``outer concatenation'' to have a two-dimensional array of the concatenated vectors of $\mathbf{a}_i$ and $\mathbf{b}_j$ (i.e., a third-order tensor). 
Specifically, by replicating $\mathbf{A}$ horizontally $J$ times and $\mathbf{B}$ vertically $I$ times, we have two ${I\times J\times D}$ tensors and concatenate them as an ${I\times J\times 2D}$ tensor. The tensor is then input to a Fully Convolutional Network (FCN) which functions as an attention model. In this paper, we utilize U-Net as an FCN.
\par
\begin{figure}[t]
\centering
\includegraphics[width=1.\columnwidth]{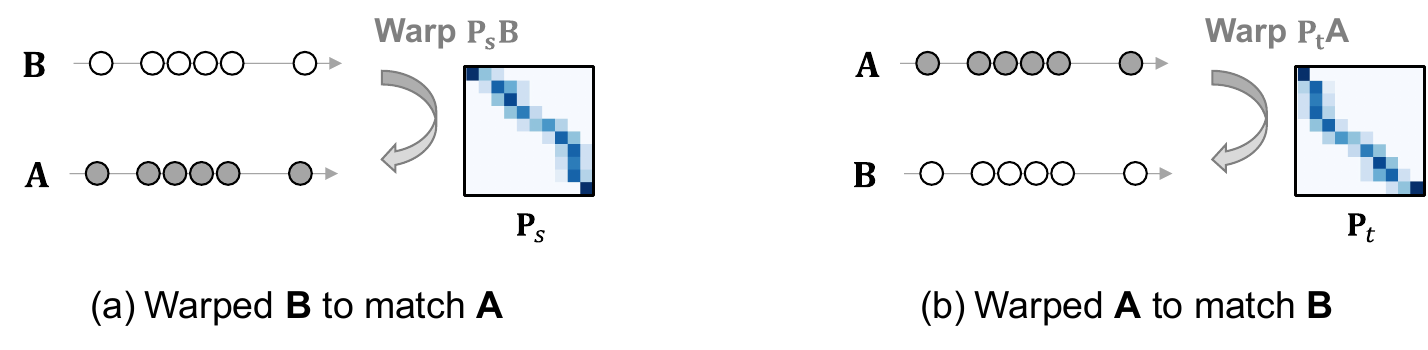}
\caption{The time warping by the attention weight matrix.}
\label{fig:warping}
\end{figure}
Before outputting an attention weight matrix $\mathbf{P}_s$, a row-wise softmax operation is applied to the output of the FCN, so that the sum of the values in the rows becomes 1.
This operation is important for using $\mathbf{P}_s$ as the soft-correspondence, as shown in Fig.~\ref{fig:overview}~(c). Consequently, the attention weight matrix $\mathbf{P}_s$ is used for warping of $\mathbf{B}$, as shown in Fig.~\ref{fig:warping}~(a). The time warping of $\mathbf{B}$ is simply given by the matrix product $\mathbf{P}_s\mathbf{B}$ and expected to be similar to  $\mathbf{A}$. In a similar way, we also have another matrix $\mathbf{P}_t$, which warps $\mathbf{A}$ to be $\mathbf{P}_t\mathbf{A}\sim\mathbf{B}$, as shown in Fig.~\ref{fig:warping}~(b). The matrix $\mathbf{P}_t$ is given by first transposing the output of FCN and then applying the row-wise softmax operation.
\par
As clarified above, the matrix $\mathbf{P}_s$ (and $\mathbf{P}_t$) is used as an attention for controlling the time series $\mathbf{B}$ ($\mathbf{A}$) to be similar to $\mathbf{A}$ ($\mathbf{B}$). The bipartite attention module drives the matrices $\mathbf{P}_s$ and $\mathbf{P}_t$ at the same time by utilizing two-dimensional nature of the outer-concatenated representation of $\mathbf{A}$ and $\mathbf{B}$. In other words, $\mathbf{A}$ is used to attend individual elements of $\mathbf{B}$ and vice versa. This mutual attention is analogous to the cost matrix of the so-called bipartite matching problem.
We, therefore, call our special attention scheme bipartite attention and differentiate from popular attention schemes such as additive attention~\cite{attn:add}
and dot-product attention~\cite{attn:dot_product,attn:scaled_dot_product}.\par
Since we use an FCN (U-Net) in the bipartite attention module, the proposed method, theoretically, can handle time series samples with variable lengths. Namely, the lengths $I$ and $J$ can be different among samples. In the later experiments, however, we use a fixed-length time series by following the traditional experimental setup of the comparative methods (such as DDTW and PSN). This fixed-length condition also allows efficient batch-based training.  
\subsection{Learning attention model with contrastive loss \label{subsec:metric_learning}}

\begin{figure}[t]
\centering
\includegraphics[width=1.\columnwidth]{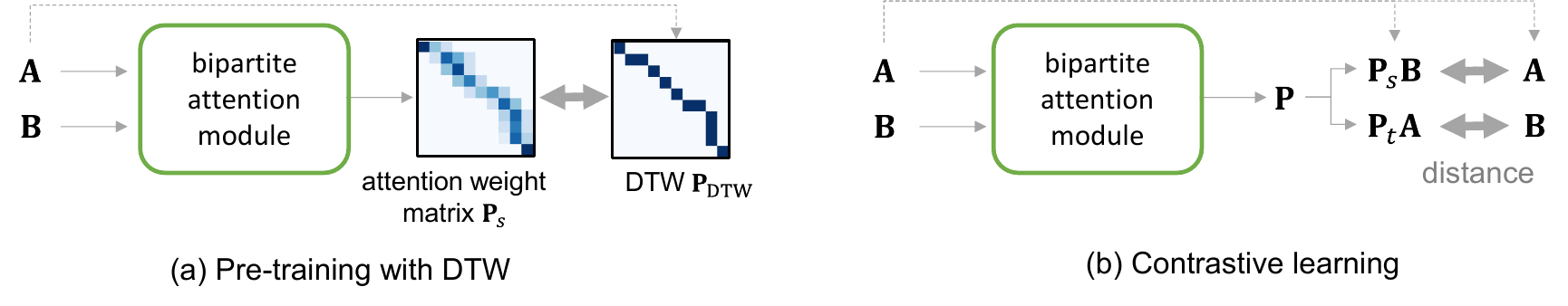}
\caption{The bipartite attention module is optimized in a two-step manner. In the first step, The module is pre-trained to mimic DTW, and in the second step, the module is optimized by contrastive training.}
\label{fig:training_step}
\end{figure}

To achieve time warping with sufficient time distortion invariance and discriminative power for a specific task, we learn the bipartite attention model with the following {\em dual contrastive loss}:
\begin{equation}
\mathcal{L}(\mathbf{A}, \mathbf{B}) =
\mathcal{L}_s(\mathbf{A}, \mathbf{B}) + \mathcal{L}_t(\mathbf{A}, \mathbf{B}).
\end{equation}
Both $\mathcal{L}_s$ and $\mathcal{L}_t$ are formulated as a
 contrastive loss~\cite{HadsellCL06} specialized for the proposed method. More specifically,  $\mathcal{L}_s$ is formulated as 
\begin{equation}
\mathcal{L}_s(\mathbf{A}, \mathbf{B})=
\begin{cases}
 \frac{1}{ID} \|\mathbf{A}-\mathbf{P}_s\mathbf{B}\|_\mathrm{F}^2&\text{if a same-class pair},\\
\max\left(0,\tau- \frac{1}{ID} \|\mathbf{A}-\mathbf{P}_s\mathbf{B}\|_\mathrm{F}^2\right)&\text{otherwise}, 
\end{cases}
\label{e:contrastive_A}
\end{equation}
where $\tau$ is the hyper-parameter for margin. $\|\cdot\|_\mathrm{F}$ denotes the Frobenius norm.
If $\mathbf{A}$ and $\mathbf{B}$ are a same-class pair, the distance between the input $\mathbf{A}$ and the warped input $\mathbf{P}_s\mathbf{B}$ is minimized. If not, their distance is optimized to be larger than $\tau$. The other loss $\mathcal{L}_t$ is defined by using $\|\mathbf{B}-\mathbf{P}_t\mathbf{A}\|_\mathrm{F}$ and $J$ (the length of $\mathbf{B}$), instead. Fig.~\ref{fig:training_step}~(b) summarizes the above process to train the bipartite attention module. 
\par
It should be emphasized that the above contrastive learning has a clear advantage over the standard DTW. The objective of the standard DTW is to minimize the distance between two time series regardless of whether they belong to the same class or not. Therefore, the standard DTW often underestimates the distance for different-class pairs. In contrast, the proposed method considers their classes and therefore can have appropriate time distortion invariance and discriminability at the same time.



\subsection{Pre-training with the standard DTW \label{subsec:pre-training}}
The proposed deep attentive time warping has much more warping flexibility than the standard DTW. As reviewed in \ref{subsec:dtw}, several constraints, such as monotonicity and continuity, are imposed to control the warping path in the standard DTW. Since the proposed method does not have such constraints, it has higher flexibility. However, of course, too much flexibility is not appropriate for many applications.\par
We, therefore, introduce a pre-training phase with the standard DTW so that the proposed method can mimic the DTW before starting its main training phase. 
Specifically, as shown in Fig.~\ref{fig:training_step}~(a), 
we prepare a binary matrix $\mathbf{P}_\mathrm{DTW}$ showing the DTW path between $\mathbf{A}$ and $\mathbf{B}$, then 
pre-train the FCN to minimize the following loss function:
\begin{equation}
\mathcal{L}_\mathrm{pre}(\mathbf{P}_s, \mathbf{P}_\mathrm{DTW})=
 \frac{1}{IJ} \|\mathbf{P}_{s}-\mathbf{P}_\mathrm{DTW}\|_\mathrm{F}^2.
\label{e:pre-training}
\end{equation}\par
After the above pre-training phase, the attention model is further trained using the contrastive loss, as described in Section~\ref{subsec:metric_learning}. By this two-step training scheme, the proposed method can avoid excessive warping flexibility, while keeping more flexibility than the standard DTW. We confirm the positive effect of pre-training through ablation studies in later experiments.
\section{Preliminary Experiments in Stand-Alone Scenario \label{sec:exp1}}
We conducted the experimental evaluation of the proposed deep attentive time warping in the stand-alone scenario. As shown in Fig.~\ref{fig:scenarios}~(a), this scenario uses the proposed method for calculating a distance between two time series and then the distance can be used in, for example, a nearest-neighbor classifier.
First, we conduct qualitative evaluations and show the behavior of the proposed method on the online handwritten character dataset, called Unipen, as a simple example. Next, we conduct quantitative evaluations and show the proposed method has a better trade-off between robustness against time distortion and discriminative power than DTW on 52 datasets of the famous UCR Archive. \par
Note that the experiments in this section mainly aim to confirm the time warping ability of the proposed method, and thus comparative study will be made with rather traditional DTW methods. The comparisons with state-of-the-art learnable time warping methods will be shown in the next section.
\subsection{Qualitative evaluations using online handwritten samples \label{subsec:qualitative_analysis}}
\subsubsection{Unipen Dataset}
Unipen~\cite{Unipen} is comprised of several subsets and we used the most popular ones, Unipen 1a (digits, 10 classes, 7,562 samples in total), Unipen 1b (uppercase alphabet, 26 classes, 6,039 samples), and Unipen 1c (lowercase alphabet, 26 classes, 10,712 samples), for the evaluation. Each sample is a sequence of 2D pen-tip coordinate vectors. For the detailed comparison with fixed-length methods (such as SVM, 1D-CNN, and Siamese), linear resampling is performed on each sample so that their temporal length was 50. The 2D coordinates were normalized to the range $[-1, 1]$. For each class, 200 samples were randomly selected for validation, and other 200 for test. All the remaining samples were used for training.\par
\subsubsection{Implementation details \label{subsec:ex1_implementation_details}}
The network architecture in the bipartite attention module follows the original U-net~\cite{U-Net}, except for an additional batch normalization layer after each convolutional layer.
The learning rate was set to $0.0001$, and Adam~\cite{ADAM15} was used as the optimizer. Before pre-training, the network weights were initialized by He initialization~\cite{HeZRS15}. The batch size was set to $512$.
During training, same-class and different-class pairs were loaded in a ratio of $1:2$. The hyper-parameter $\tau$ in the contrastive loss was set to $1$. 
The maximum iterations for pre-training of Fig.~\ref{fig:training_step}~(a) and the main contrastive training of (b) were set at $1,000$ and $10,000$, respectively; and the best model (i.e., the best iteration number) was chosen
by the evaluation with the validation set. \par
For quantitative evaluation, we conducted a classification experiment using the distance by the proposed method. For each test sample, its distances to all training samples were calculated by the proposed method and the $k$-nearest neighbor classification was 
performed to determine 
its class label at$k=1$. As the distance 
between $\mathbf{A}$ and $\mathbf{B}$, we use the following ``symmetric'' distance: 
\begin{equation}
d(\mathbf{A}, \mathbf{B})=
 \frac{1}{ID}\|\mathbf{A}-\mathbf{P}_s\mathbf{B}\|_\mathrm{F}^2 +  \frac{1}{JD}\|\mathbf{B}-\mathbf{P}_t\mathbf{A}\|_\mathrm{F}^2.\label{e:distance}
\end{equation}\par
The proposed method achieved 
99.0, 98.0, and 95.5\% classification accuracies for 
Unipen 1a, 1b, and 1c, respectively, whereas the standard DTW achieved 98.4, 96.0, and 94.1\%. This proves the proposed method achieved sufficient accuracies and, therefore, the following qualitative evaluation results are reliable enough.
\subsubsection{Qualitative evaluation results\label{subsubsec:unipen}}

\begin{figure}[t]
\centering
\subfloat[][Unipen1b]{
\includegraphics[width=.3\columnwidth]{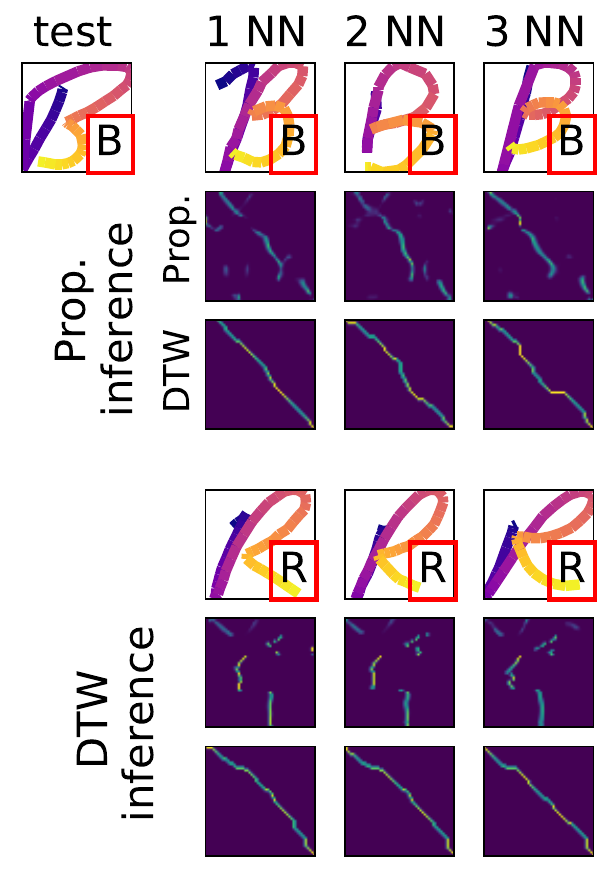}
\hspace{3mm}
\includegraphics[width=.3\columnwidth]{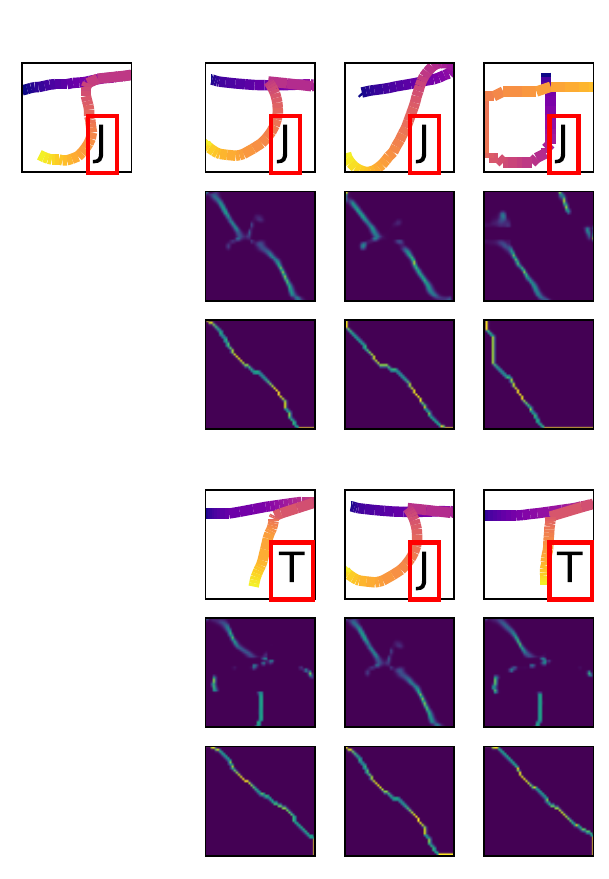}
\hspace{3mm}
\includegraphics[width=.3\columnwidth]{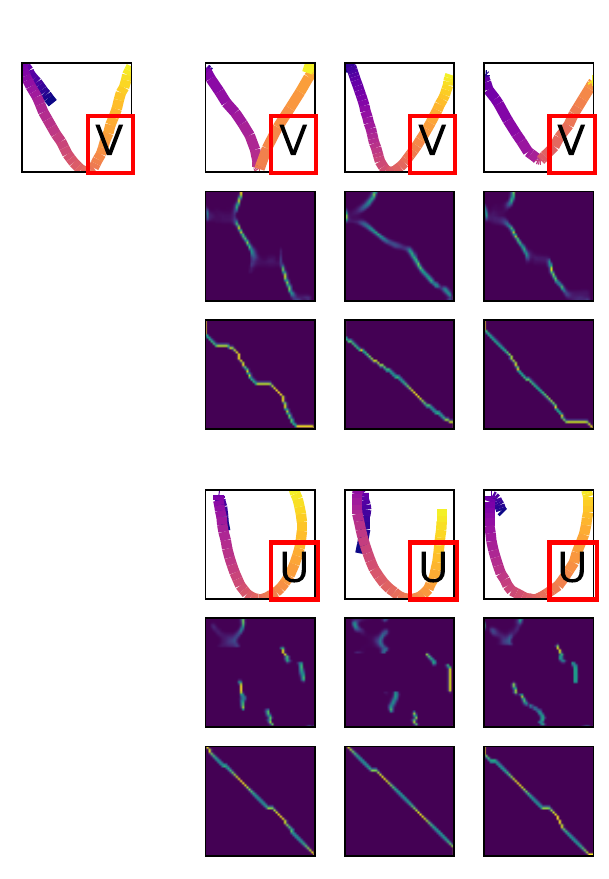}
}\\
\subfloat[][Unipen1c]{
\includegraphics[width=.3\columnwidth]{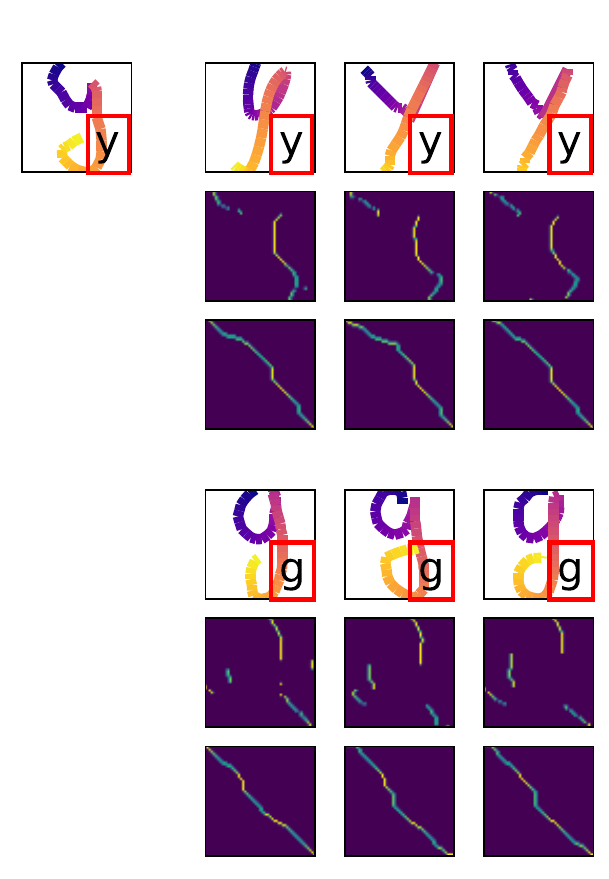}
\hspace{3mm}
\includegraphics[width=.3\columnwidth]{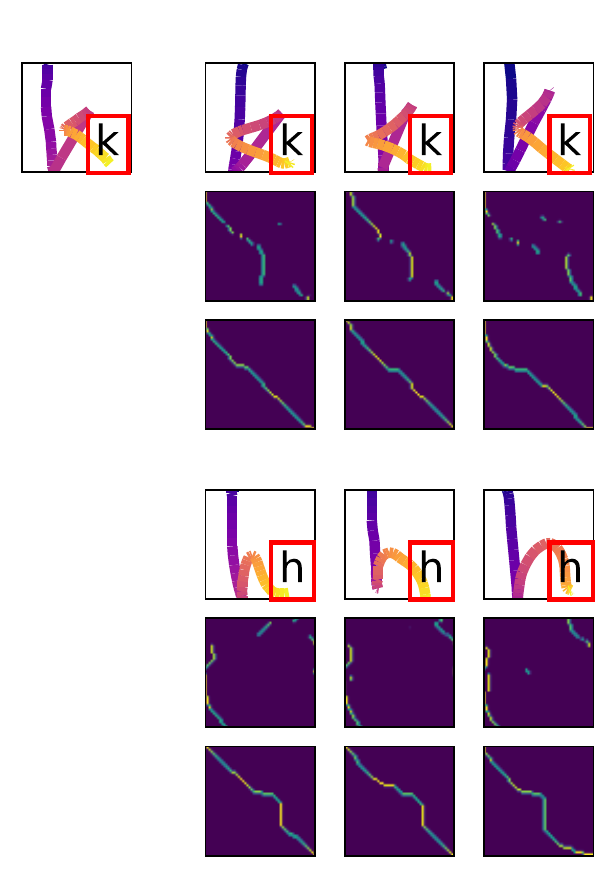}
\hspace{3mm}
\includegraphics[width=.3\columnwidth]{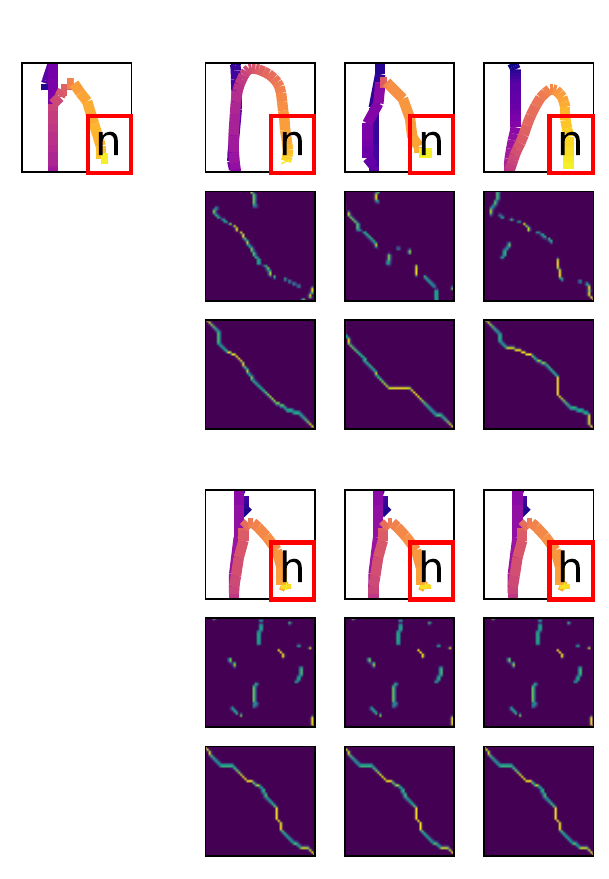}
}
\caption{A visualization of the matching path of the improved samples compared with DTW. The character in the red box shows the ground truth.}
\label{fig:visualization_path}
\end{figure}

Figs.~\ref{fig:visualization_path}~(a) and (b) show the results on three test samples of Unipen 1b and 1c, respectively. Those test samples are correctly classified by the proposed method and not by DTW. 
For each test sample, the top three nearest neighbors by distance of the proposed method and those by the DTW distance are shown. The attention weight matrices
and the DTW matching paths are also shown.\par

In Fig.~\ref{fig:visualization_path}~(a), the proposed method classifies the test sample as `\texttt{B}' correctly by attention matrices that resemble the DTW path. 
In contrast, DTW classifies this  `\texttt{B}'  as `\texttt{R}' incorrectly. It should be emphasized that the proposed method provided an almost meaningless attention matrix between `\texttt{B}' and `\texttt{R},' {\em intentionally}. This is because the proposed method tries to differentiate them as an expected effect of its contrastive learning. Similar attention matrices are found in other cases. Since DTW has no such function, it always gives smooth correspondence and gets a small distance that causes misclassification. \par

The third nearest neighbor of `\texttt{J}' in the middle column of Fig.~\ref{fig:visualization_path}~(a) shows another benefit of the proposed method. This `\texttt{J}' shows a different stroke order. Since the proposed method does not have a strict monotonicity constraint, its attention map deals with the stroke order variation.
From these results, we can observe that the proposed method has an appropriate time warping flexibility that realizes both sufficient time distortion invariance and discriminability.  

\begin{figure}[t]
\centering
\includegraphics[width=.99\columnwidth]{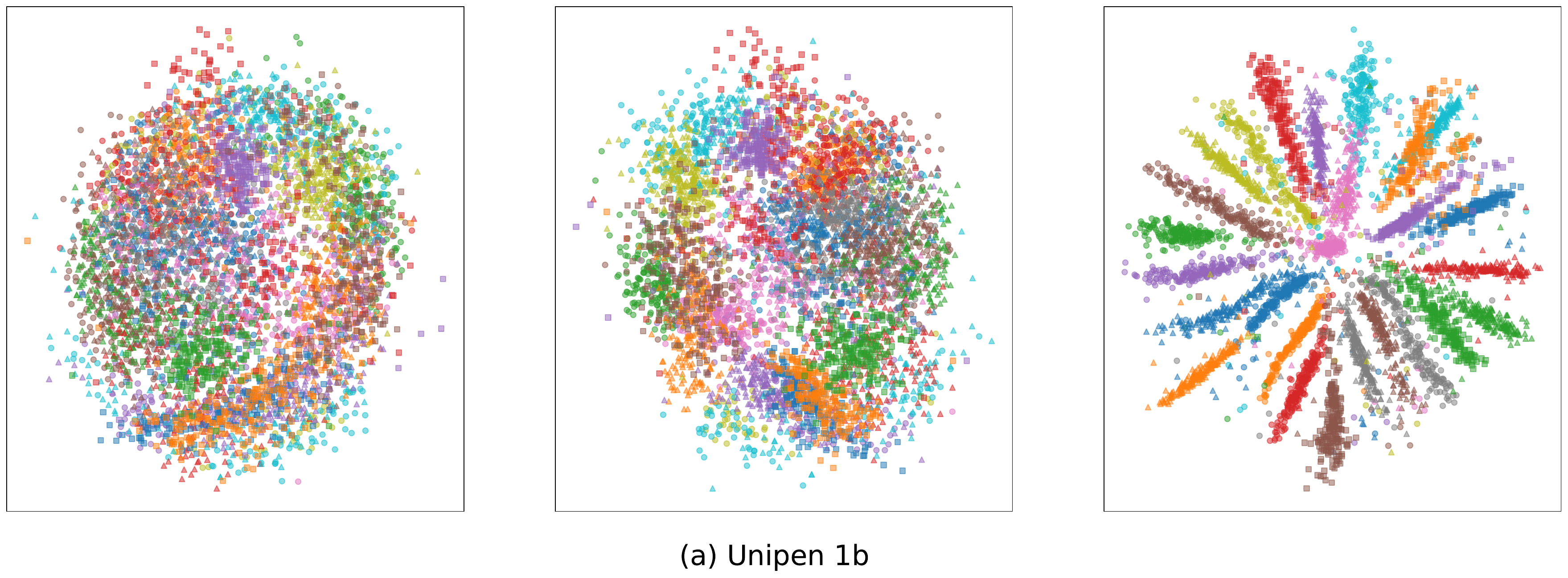}\\ \vspace{3mm}
\includegraphics[width=.99\columnwidth]{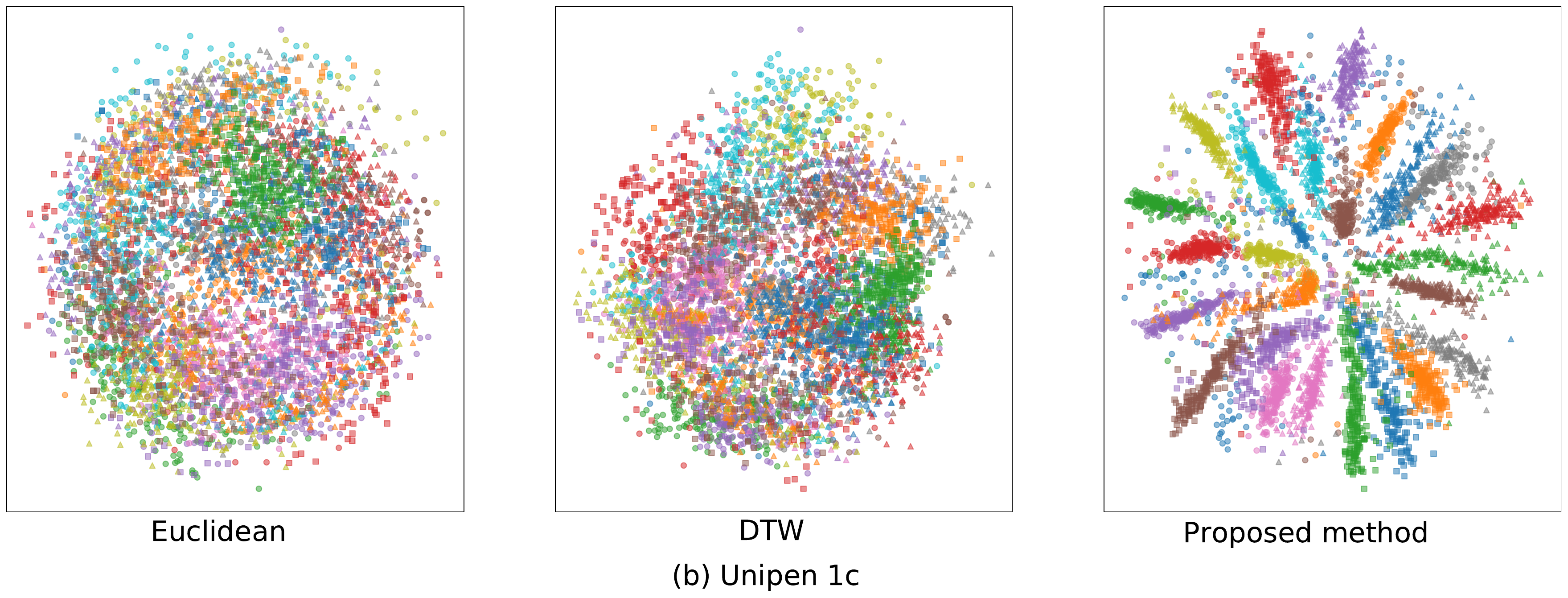}\\
\includegraphics[width=.99\columnwidth]{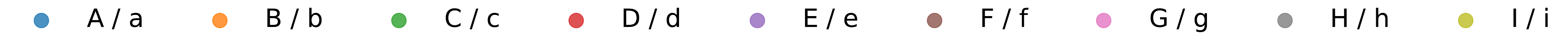}\\ \vspace{-3mm}
\includegraphics[width=.99\columnwidth]{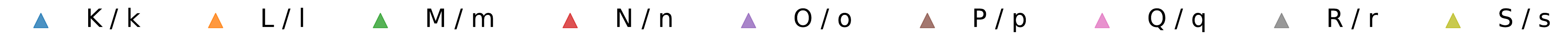}\\ \vspace{-3mm} 
\hspace{-12mm}
\includegraphics[width=.9\columnwidth]{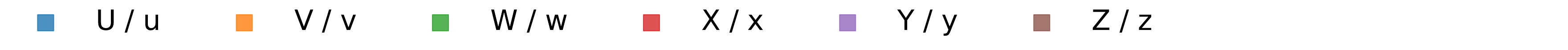}

\caption{The visualization of the test samples by MDS on (a)~Unipen 1b and (b)~1c.}
\label{fig:MDSe_unipen}
\end{figure}

Fig.~\ref{fig:MDSe_unipen} shows the distribution of test samples by the multi-dimensional scaling (MDS). Three distance metrics, Euclidean, DTW, and the proposed method, are used for these MDS visualizations. For the proposed method, the distance between a pair of test samples $\mathbf{A}$ and $\mathbf{B}$ is evaluated by (\ref{e:distance}). \par
These distributions prove that the distance by DTW is more discriminative than Euclidean, and the distance by the proposed method is far more discriminative than DTW. For example, the overlap between `\texttt{U}'
and `\texttt{V}' in Unipen 1b by the DTW distance disappears in the proposed method. The contrastive metric learning in the proposed method realizes this discriminability, as expected.\par

\begin{table}[t]
\centering
\caption{Error rates (\%) between confusing classes (Unipen). Error rates in red indicate the least rate of each case.}\vspace{-2mm}
\label{tab:unipen_error}
\scalebox{0.9}[0.9]{
\begin{tabular*}{\linewidth}{@{\extracolsep{\fill}}lrrrrr}
\toprule
& \multicolumn{2}{c}{Unipen 1b} & \multicolumn{3}{c}{Unipen 1c} \\
\cmidrule(lr){2-3}
\cmidrule(lr){4-6}
~Method & `\texttt{J}' vs. `\texttt{T}' & `\texttt{U}' vs. `\texttt{V}' & `\texttt{g}' vs. `\texttt{y}' & `\texttt{h}' vs. `\texttt{k}' & `\texttt{h}' vs. `\texttt{n}' \\
\midrule
~ours & \textcolor{red}{3.5} & \textcolor{red}{6.5} & \textcolor{red}{3.0} & \textcolor{red}{3.0} & \textcolor{red}{6.5} \\
~DTW & 14.0 & 11.5 & 12.0 & 8.0 & 10.0 \\
\bottomrule
\end{tabular*}
}
\end{table}

\begin{figure}[t]
\centering
\subfloat[`\texttt{J}' vs. `\texttt{T}']{
\includegraphics[width=.23\columnwidth]{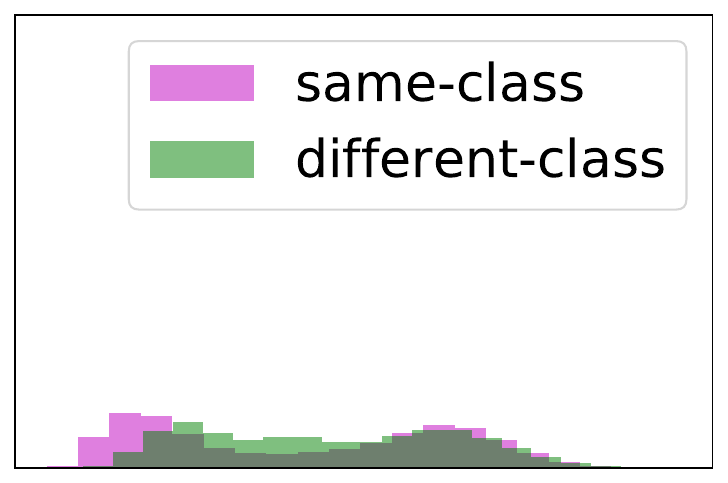}
\includegraphics[width=.23\columnwidth]{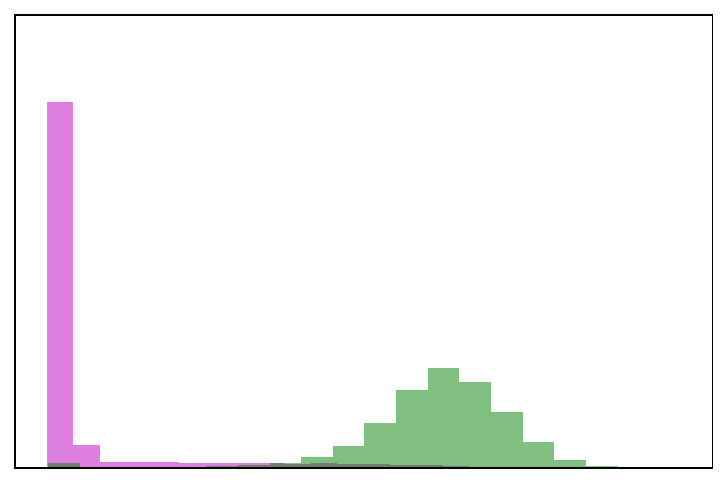}
\label{subfig:JvsT}}
\hfill
\subfloat[`\texttt{U}' vs. `\texttt{V}']{
\includegraphics[width=.23\columnwidth]{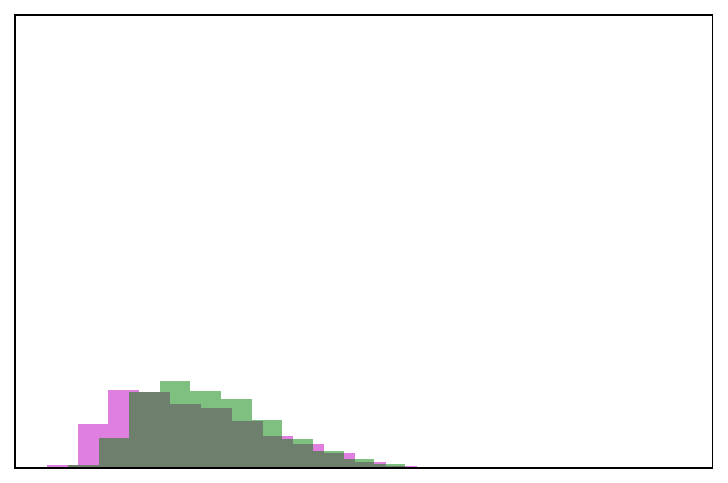}
\includegraphics[width=.23\columnwidth]{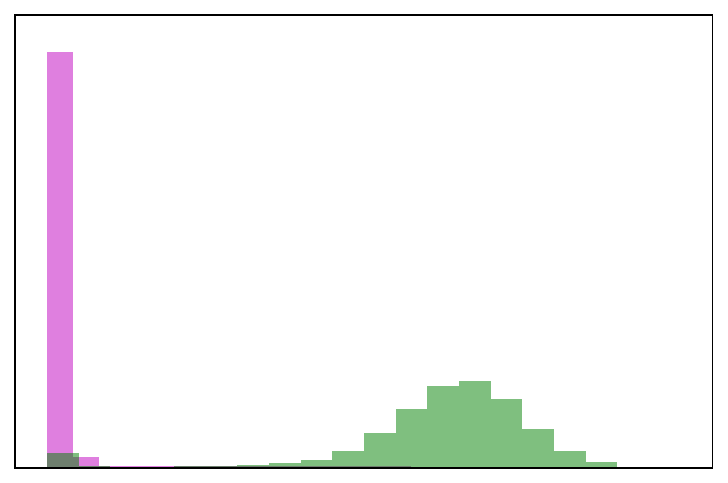}
\label{subfig:UvsV}}\\[-3mm]
\caption{Distance histograms for two confusing class pairs, `\texttt{J}' vs. `\texttt{T}' and `\texttt{U}' vs. `\texttt{V},' of Unipen 1b. For each pair,  
 the left histogram is about DTW distance and the right is the proposed method.}
\label{fig:unipen1b_hist}
\end{figure}

Table~\ref{tab:unipen_error} shows the error rates for binary classifications between ambiguous class pairs in Unipen 1b and 1c. Fig.~\ref{fig:unipen1b_hist} shows the normalized histograms of the distances by DTW and the proposed method for the ambiguous class pairs in Unipen 1b. The histogram of DTW shows a large overlap between the same and different-class pairs, whereas the proposed method does not.
These results also prove the sufficient discriminative power of the proposed method. \par
\subsection{Quantitative analysis on UCR dataset\label{subsec:quntitative_analysis}}
\subsubsection{UCR Dataset}
\label{subsubs:dataset_ucr}
The University of California Riverside (UCR) Time Series Classification Archive (2015 edition)~\cite{UCRArchive} is a famous benchmark that is comprised of 85 different univariate time series datasets. Among them, we selected 52 datasets satisfying the following two conditions. The first condition requires that more than 100 training samples are available. The second requires that the sample length ($I$ and $J$) should be less than 1,000. The proposed method can, theoretically, deal with any sample length (even longer than 1,000); however, in practice, too long of samples cause memory issues (like other trainable time warping methods). In each dataset, all the samples are already regulated to have the same length. 
UCR prepares a training sample set and a test sample set for each dataset. Among the training samples, 90\% is used for training and 10\% for validation. All time series were standardized for each channel to have a mean of zero and a variance of one.\par
\begin{table}[p!]
\color{red}
\centering
\caption{Error rates by the proposed method (ours) and the comparative methods. Error rates in \textcolor{red}{red} and \textcolor{blue}{blue} indicate the least and the second least rates, respectively.
If an error rate of a comparative method is printed in \textbf{bold}, the proposed method was superior to the comparative method with statistical significance at the 5\% level by McNemar's test. If \textit{italic}, the proposed method is inferior with significance at the 5\% level.}

\color{black}
\label{tab:UCR}
\scalebox{0.6}[0.53]{
\begin{tabular}{lrrrrrrrrr}
\toprule
Dataset                        & Train & Test & Class & Length & DTW                               & w-DTW                              & s-DTW                              & ours                      & w/o pre-train.                    \\ \midrule
Adiac                          & 390   & 391  & 37    & 176    & 39.64                             & $\textcolor{blue}{38.87}$          & 41.94                              & 42.46                     & $\textit{\textcolor{red}{28.13}}$ \\
ChlorineConcentration          & 467   & 3840 & 3     & 166    & $\textbf{35.16}$                  & $\textbf{35.16}$                   & $\textbf{35.42}$                   & $\textcolor{blue}{18.72}$ & $\textcolor{red}{18.41}$          \\
Computers                      & 250   & 250  & 2     & 720    & 37.60                             & 43.60                              & $\textcolor{red}{34.80}$           & $\textcolor{blue}{37.20}$ & 40.00                             \\
CricketX                       & 390   & 390  & 12    & 300    & $\textcolor{blue}{24.36}$         & 24.87                              & $\textit{\textcolor{red}{22.82}}$  & 28.21                     & 25.90                             \\
CricketY                       & 390   & 390  & 12    & 300    & $\textcolor{blue}{25.13}$         & $\textcolor{blue}{25.13}$          & 25.90                              & $\textcolor{red}{24.87}$  & 29.74                             \\
CricketZ                       & 390   & 390  & 12    & 300    & 24.36                             & $\textbf{29.49}$                   & $\textcolor{red}{21.79}$           & $\textcolor{blue}{23.33}$ & $\textbf{28.97}$                  \\
DistalPhalanxOutlineAgeGroup   & 400   & 139  & 3     & 80     & $\textit{\textcolor{red}{23.02}}$ & $\textit{\textcolor{red}{23.02}}$  & 23.74                              & 33.81                     & 30.22                             \\
DistalPhalanxOutlineCorrect    & 600   & 276  & 2     & 80     & 28.26                             & 27.54                              & $\textcolor{red}{23.91}$           & $\textcolor{blue}{25.36}$ & 26.81                             \\
DistalPhalanxTW                & 400   & 139  & 6     & 80     & 41.01                             & 41.01                              & 39.57                              & $\textcolor{red}{38.13}$  & $\textcolor{blue}{38.85}$         \\
Earthquakes                    & 322   & 139  & 2     & 512    & 28.78                             & 31.65                              & $\textcolor{red}{22.30}$           & 25.90                     & $\textcolor{blue}{25.18}$         \\
ECG200                         & 100   & 100  & 2     & 96     & $\textbf{23.00}$                  & $\textbf{23.00}$                   & 18.00                              & $\textcolor{red}{9.00}$   & $\textcolor{blue}{13.00}$         \\
ECG5000                        & 500   & 4500 & 5     & 140    & $\textit{7.56}$                   & $\textit{7.49}$                    & $\textit{\textcolor{blue}{7.27}}$  & 9.40                      & $\textit{\textcolor{red}{7.00}}$  \\
ElectricDevices                & 8926  & 7711 & 7     & 96     & 40.50                             & $\textit{\textcolor{red}{38.28}}$  & $\textit{\textcolor{blue}{39.37}}$ & 41.33                     & $\textit{40.37}$                  \\
FaceAll                        & 560   & 1690 & 14    & 131    & $\textbf{19.23}$                  & $\textbf{18.28}$                   & $\textbf{20.24}$                   & $\textcolor{red}{16.51}$  & $\textcolor{blue}{17.34}$         \\
FacesUCR                       & 200   & 2050 & 14    & 131    & $\textbf{9.51}$                   & $\textbf{8.39}$                    & $\textbf{8.63}$                    & $\textcolor{red}{4.44}$   & $\textbf{\textcolor{blue}{5.90}}$ \\
FiftyWords                     & 450   & 455  & 50    & 270    & $\textbf{30.55}$                  & $\textbf{27.69}$                   & 20.44                              & $\textcolor{red}{18.90}$  & $\textcolor{blue}{19.12}$         \\
Fish                           & 175   & 175  & 7     & 463    & $\textbf{17.71}$                  & $\textbf{16.57}$                   & 13.71                              & $\textcolor{blue}{8.00}$  & $\textcolor{red}{5.71}$           \\
FordA                          & 3601  & 1320 & 2     & 500    & $\textbf{44.70}$                  & $\textbf{32.20}$                   & $\textbf{39.02}$                   & $\textcolor{red}{8.48}$   & $\textcolor{blue}{10.00}$         \\
FordB                          & 3636  & 810  & 2     & 500    & $\textbf{38.02}$                  & $\textbf{37.90}$                   & $\textbf{41.98}$                   & $\textcolor{red}{19.14}$  & $\textcolor{blue}{19.51}$         \\
Ham                            & 109   & 105  & 2     & 431    & $\textbf{52.38}$                  & $\textbf{52.38}$                   & $\textbf{41.90}$                   & $\textcolor{red}{25.71}$  & $\textcolor{blue}{30.48}$         \\
InsectWingbeatSound            & 220   & 1980 & 11    & 256    & $\textbf{63.84}$                  & $\textcolor{red}{42.98}$           & $\textcolor{blue}{44.09}$          & 44.95                     & 47.63                             \\
LargeKitchenAppliances         & 375   & 375  & 3     & 720    & $\textit{26.40}$                  & $\textit{\textcolor{blue}{26.13}}$ & $\textit{\textcolor{red}{18.40}}$  & 43.20                     & 45.60                             \\
MedicalImages                  & 381   & 760  & 10    & 99     & $\textit{26.32}$                  & $\textit{\textcolor{blue}{25.26}}$ & $\textit{\textcolor{red}{24.87}}$  & 31.71                     & 34.61                             \\
MiddlePhalanxOutlineAgeGroup   & 400   & 154  & 3     & 80     & $\textcolor{blue}{50.00}$         & $\textcolor{blue}{50.00}$          & 54.55                              & $\textcolor{red}{48.70}$  & 55.19                             \\
MiddlePhalanxOutlineCorrect    & 600   & 291  & 2     & 80     & $\textbf{30.24}$                  & $\textbf{30.24}$                   & $\textbf{27.84}$                   & $\textcolor{red}{17.18}$  & $\textcolor{blue}{21.31}$         \\
MiddlePhalanxTW                & 399   & 154  & 6     & 80     & 49.35                             & 49.35                              & $\textcolor{blue}{48.05}$          & 50.00                     & $\textcolor{red}{45.45}$          \\
NonInvasiveFetalECGThorax1     & 1800  & 1965 & 42    & 750    & 20.97                             & $\textit{\textcolor{red}{18.47}}$  & $\textit{\textcolor{blue}{19.80}}$ & 22.14                     & $\textbf{49.26}$                  \\
NonInvasiveFetalECGThorax2     & 1800  & 1965 & 42    & 750    & $\textit{13.54}$                  & $\textit{\textcolor{blue}{12.26}}$ & $\textit{\textcolor{red}{11.45}}$  & 18.73                     & $\textit{15.78}$                  \\
OSULeaf                        & 200   & 242  & 6     & 427    & $\textbf{40.50}$                  & $\textbf{43.39}$                   & $\textbf{35.12}$                   & $\textcolor{blue}{25.21}$ & $\textit{\textcolor{red}{17.77}}$ \\
PhalangesOutlinesCorrect       & 1800  & 858  & 2     & 80     & $\textbf{27.16}$                  & $\textbf{27.16}$                   & $\textbf{28.09}$                   & $\textcolor{blue}{21.68}$ & $\textcolor{red}{19.46}$          \\
Plane                          & 105   & 105  & 7     & 144    & $\textcolor{red}{0.00}$           & $\textcolor{red}{0.00}$            & $\textcolor{red}{0.00}$            & $\textcolor{red}{0.00}$   & $\textcolor{red}{0.00}$           \\
ProximalPhalanxOutlineAgeGroup & 400   & 205  & 3     & 80     & 19.51                             & 19.51                              & 20.98                              & $\textcolor{red}{17.56}$  & $\textcolor{blue}{19.02}$         \\
ProximalPhalanxOutlineCorrect  & 600   & 291  & 2     & 80     & $\textbf{21.65}$                  & $\textbf{20.96}$                   & $\textbf{23.02}$                   & $\textcolor{red}{8.59}$   & $\textcolor{blue}{9.97}$          \\
ProximalPhalanxTW              & 400   & 205  & 6     & 80     & $\textcolor{blue}{24.39}$         & $\textcolor{blue}{24.39}$          & 25.37                              & $\textcolor{blue}{24.39}$ & $\textcolor{red}{22.44}$          \\
RefrigerationDevices           & 375   & 375  & 3     & 720    & $\textbf{54.67}$                  & $\textbf{57.87}$                   & $\textbf{\textcolor{blue}{54.13}}$ & $\textcolor{red}{44.27}$  & $\textbf{67.20}$                  \\
ScreenType                     & 375   & 375  & 3     & 720    & $\textcolor{blue}{60.80}$         & 64.27                              & 65.60                              & 64.27                     & $\textcolor{red}{59.20}$          \\
ShapesAll                      & 600   & 600  & 60    & 512    & $\textbf{23.00}$                  & $\textbf{23.67}$                   & $\textbf{17.33}$                   & $\textcolor{red}{13.67}$  & $\textcolor{red}{13.67}$          \\
SmallKitchenAppliances         & 375   & 375  & 3     & 720    & $\textit{\textcolor{red}{31.73}}$ & 41.33                              & $\textit{\textcolor{blue}{34.67}}$ & 45.07                     & 48.53                             \\
Strawberry                     & 613   & 370  & 2     & 235    & 5.95                              & 5.95                               & 5.68                               & $\textcolor{blue}{5.14}$  & $\textcolor{red}{4.32}$           \\
SwedishLeaf                    & 500   & 625  & 15    & 128    & $\textbf{20.80}$                  & $\textbf{15.20}$                   & $\textbf{17.44}$                   & $\textcolor{red}{7.36}$   & $\textbf{\textcolor{blue}{9.76}}$ \\
SyntheticControl               & 300   & 300  & 6     & 60     & 0.67                              & 0.67                               & 0.67                               & $\textcolor{blue}{0.33}$  & $\textcolor{red}{0.00}$           \\
Trace                          & 100   & 100  & 4     & 275    & $\textcolor{red}{0.00}$           & 5.00                               & 1.00                               & $\textcolor{red}{0.00}$   & $\textcolor{red}{0.00}$           \\
TwoPatterns                    & 1000  & 4000 & 4     & 128    & $\textcolor{red}{0.00}$           & $\textbf{0.68}$                    & $\textcolor{red}{0.00}$            & $\textcolor{red}{0.00}$   & $\textcolor{red}{0.00}$           \\
UWaveGestureLibraryAll         & 896   & 3582 & 8     & 945    & $\textit{5.28}$                   & $\textit{\textcolor{blue}{4.75}}$  & $\textit{\textcolor{red}{4.16}}$   & 9.24                      & $\textbf{88.11}$                  \\
UWaveGestureLibraryX           & 896   & 3582 & 8     & 315    & $\textbf{26.91}$                  & $\textbf{24.46}$                   & $\textcolor{blue}{22.59}$          & $\textcolor{red}{22.03}$  & $\textbf{24.23}$                  \\
UWaveGestureLibraryY           & 896   & 3582 & 8     & 315    & $\textbf{36.21}$                  & $\textcolor{blue}{31.88}$          & $\textit{\textcolor{red}{29.93}}$  & 32.94                     & 33.56                             \\
UWaveGestureLibraryZ           & 896   & 3582 & 8     & 315    & $\textbf{34.06}$                  & $\textbf{\textcolor{blue}{32.80}}$ & $\textbf{32.89}$                   & $\textcolor{red}{29.09}$  & $\textbf{34.34}$                  \\
Wafer                          & 1000  & 6164 & 2     & 152    & $\textbf{2.01}$                   & $\textbf{0.44}$                    & $\textbf{0.71}$                    & $\textcolor{red}{0.18}$   & $\textcolor{blue}{0.23}$          \\
WordSynonyms                   & 267   & 638  & 25    & 270    & 34.64                             & $\textcolor{blue}{29.47}$          & $\textit{\textcolor{red}{23.35}}$  & 32.76                     & $\textbf{43.10}$                  \\
Worms                          & 181   & 77   & 5     & 900    & 49.35                             & $\textcolor{blue}{46.75}$          & $\textcolor{blue}{46.75}$          & $\textcolor{red}{37.66}$  & $\textbf{57.14}$                  \\
WormsTwoClass                  & 181   & 77   & 2     & 900    & 42.86                             & 41.56                              & $\textcolor{red}{35.06}$           & $\textcolor{blue}{38.96}$ & 48.05                             \\
Yoga                           & 300   & 3000 & 2     & 426    & 16.37                             & $\textcolor{blue}{15.67}$          & $\textit{\textcolor{red}{14.07}}$  & 17.10                     & $\textbf{42.87}$                  \\
All                            &       &      &       &        & $\textbf{23.58}$                  & $\textbf{21.69}$                   & $\textbf{21.39}$                   & 20.21                     & $\textbf{26.96}$         \\
\toprule
\multicolumn{5}{c}{Average}                                                                                                                    &        27.88            &     27.21                 & \textcolor{blue}{25.58}                    &  \textcolor{red}{23.71}                   & 27.66                             \\
\multicolumn{5}{c}{Wins}                                                                                                                       &         5               &         5                &      \textcolor{blue}{15}                  & \textcolor{red}{23}            &           \textcolor{blue}{15}                   \\  
\toprule
\end{tabular}
}
\end{table}

\subsubsection{Implementation details\label{subsec:ucr_implementation_details}}
\label{subsubs:implementation_ucr}
The model architecture, learning rate, optimizer, hyper-parameter, number of iterations, and inference protocol are the same as \ref{subsec:ex1_implementation_details}.
The batch size was determined for the maximum memory utilization of the GPU (Tesla V100).\par

We compared the proposed method with the standard DTW (DTW)~\cite{DTW78}, window-DTW (w-DTW)~\cite{DTW78} and soft-DTW (s-DTW)~\cite{soft-DTW}. The optimal values of the hyperparameters in the comparative methods, as well as the proposed method, were chosen by the validation set\footnote{Exceptionally, the value of the hyperprameter ``window size'' in w-DTW was taken from the list of {\tt https://www.cs.ucr.edu/\textasciitilde eamonn/time\_series\_data\_2018/}.}. As noted above, 10\% of UCR training set were used as the validation set. For example, the hyperparameter $\gamma$ in s-DTW was chosen from $0.01$, $0.1$, $1$, $10$, $100$ using the validation set.
\par 
We used the 1-Nearest Neighbor (1-NN) rule as the classifier. More specifically, we used the distance given by the proposed method and then compared each test sample with all training samples. The class of the training sample with the minimum distance was considered as the classification result. We used the same 1-NN classification approach for the comparative methods.\par 

\subsubsection{Quantitative evaluation results\label{sec:quantitative-UCR}}

\begin{figure}[t]

\centering
	\subfloat[DTW vs. ours]{%
		\includegraphics[clip, width=0.49\columnwidth]{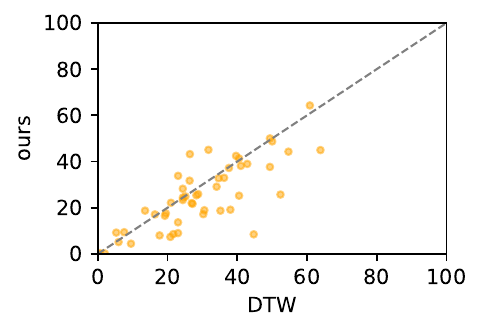}
		\label{subfig:UCR-DTWvsProp}}%
	\subfloat[ours w/o vs. w/ pre-training]{%
		\includegraphics[clip, width=0.49\columnwidth]{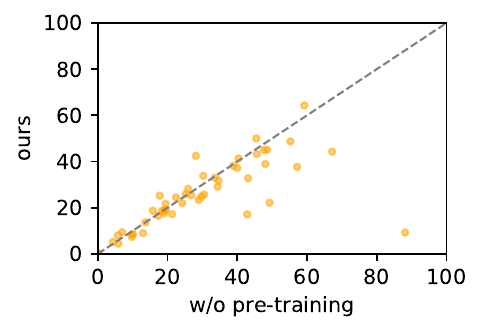}\label{subfig:UCR-Pretraining}}%
	\caption{Error rate comparison. Each point corresponds to one of the 52 datasets.}
	\label{fig:UCR_classification_result}

\end{figure}

Table~\ref{tab:UCR} shows classification error rates by the proposed method (ours), and the comparative methods, i.e., DTW, w-DTW, s-DTW, on 52 UCR2015 datasets\footnote{In UCR2018~\cite{UCRArchive2018}, we found six datasets that satisfy the same conditions as UCR2015. The experimental evaluation results on these datasets are given in Appendix~\ref{app:UCR2018}.}. As an ablation study, the performance of the proposed method without pre-training phase is also listed in this table. The error rates in red and blue indicate the least rate and the second least rate, respectively. \par 

For many datasets, the proposed deep attentive time warping achieved lower error rates than the traditional DTW methods. This fact is confirmed by the number of wins; the proposed method shows the lowest error rates for 23 among 52 datasets.\par 
Fig.~\subref*{subfig:UCR-DTWvsProp} shows a pair-wise comparison between DTW and the proposed method. Each point corresponds to one of the 52 datasets. The 36 points below the diagonal line indicate that the proposed method achieved a lower error rate than DTW for those datasets. Consequently, this figure also demonstrates a higher effectiveness of the proposed method.\par

As an ablation study, we observed the performance change by removing the pre-training phase. The accuracies of the proposed method without pre-training are shown in the rightmost column of Table~\ref{tab:UCR} (``w/o pre-train.'') and summarized in Fig.~\subref*{subfig:UCR-Pretraining} as a pairwise comparison with the method with pre-training. These results show that the positive effect of pre-training is confirmed on 36 datasets among the 52.
\par

In order to make our evaluation more reliable, 
we conducted the McNemar's test between the proposed method and each comparative method. The test results are shown in Table~\ref{tab:UCR}. If an error rate by a comparative method is printed in \textbf{bold}, the proposed method was superior to the comparative method with statistical significance at the 5\% level by McNemar's test. If \textit{italic}, the proposed method is inferior with significance at the 5\% level. \par 
From the results of McNemar's test, we can confirm the superiority of the proposed method over the comparative methods. More specifically, among 52 datasets,  the proposed method was superior to DTW, w-DTW, and s-DTW, and w/o pre-training with the statistical significance at the 5\% level on, 20, 22, 18, and 11 datasets, respectively. We can also see that inferior cases (italic) are much less than superior cases (bold). The row ``All'' in Table~\ref{tab:UCR} shows the error rates of all test samples in all 52 datasets. McNemar's test results in the ``All'' row also shows that the proposed method was superior to all the comparative methods at the 5\% level --- Precisely speaking, the proposed method was superior even at the 1\% level.\par

\begin{figure}[t]
\centering
\includegraphics[width=.4\columnwidth]{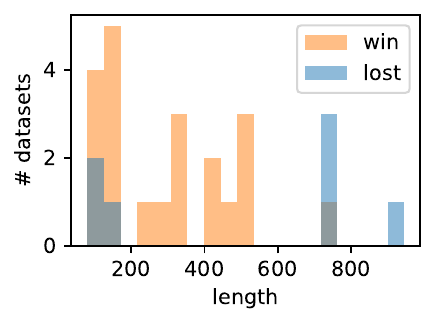}\vspace{-5mm}
\caption{Histogram of win cases and lost cases with respect to time length. Note that we only picked up the datasets that show statistical significance at the 5\% level by McNemar's test in Table~\ref{tab:UCR} for emphasizing the difference between win cases and lost cases. 
}
\label{fig:histo}
\end{figure}

In order to understand the characteristics of the proposed method, we analyzed the relationship between several dataset features (e.g., dataset size, time length, etc.) and  win-lost cases. Among these features, time length shows the most evident relationship, as shown in Fig.~\ref{fig:histo}. This figure shows the histograms of win cases and lost cases with respect to sample time length. To emphasize the difference between win cases and lost cases, we picked up the datasets that show statistical significance at the 5\% level by McNemar's test in Table~\ref{tab:UCR}. \par
The histogram suggests that the lost cases are found for the datasets with very short or very long time lengths. A possible reason for this phenomenon is the fixed network architecture of the bipartite attention module. For example, for longer samples, the network is too shallow to exchange the information between their beginning and ending parts. In future work, we can try to use different network architectures according to the time length of samples.\par

We further compare the proposed method to results reported in literature. 
We collected results that propose distance measures for a 1-NN classifier, similar to the proposed method. 
The comparative methods use a wide variety of distance measure mechanisms, including derivative based methods, Complexity-Invariant Distance~(CID)~\cite{batista2011complexity}, Derivative Transform Distance~(DTD$_C$)~\cite{G_recki_2014}, and DTW Derivative Distance~(DD$_{DTW}$)~\cite{gorecki2013using}, dictionary distance based methods, Bag of Patterns~(BOP)~\cite{lin2012rotation} and Bag of Symbolic

\begin{table}[t]
\caption{Comparison between proposed method and comparative methods on the 2015 UCR Time Series Archive datasets. The total number of datasets for each method is determined by the intersection of the datasets used by the proposed method and reported by the comparison methods.}
\label{tab:competition}
\centering
\begin{tabular*}{\linewidth}{@{\extracolsep{\fill}}lrrrr}
\toprule
 & \multicolumn{3}{c}{Proposed Method} \\
\cmidrule(lr){2-4}
Method & Wins & Losses & Ties & Total \\
\midrule
BOP~\cite{lin2012rotation} & \textbf{8} & 2 & 0 & 10 \\
BOSS~\cite{schafer2015boss} & \textbf{12} & 11 & 0 & 23 \\
CID~\cite{batista2011complexity} & \textbf{13} & 8 & 0 & 21 \\
DD$_{DTW}$~\cite{gorecki2013using} & \textbf{8} & 3 & 0 & 11 \\
DTD$_C$~\cite{G_recki_2014} & 9 & \textbf{13} & 0 & 22 \\
MSM~\cite{stefan2013move} & \textbf{8} & 3 & 0 & 11 \\
TWED~\cite{Marteau09} & \textbf{9} & 2 & 0 & 11 \\
WDTW~\cite{WDTW} & \textbf{7} & 2 & 0 & 9 \\
\bottomrule
\end{tabular*}
\vspace{2mm}
\end{table}


Table~\ref{tab:competition} lists the methods and the number times the proposed method has a higher accuracy over the comparison method (Wins), the number of times it had a lower accuracy (Loses), the number of ties, and the total number of datasets used in the comparisons. 
Note, each comparison method reports their results on different datasets within the 2015 UCR Time Series Archive. 
Therefore, we only count the datasets that are available. 
Also, since we limit the proposed method to datasets with more than 100 training patterns, we exclude the reported datasets with less than 100 training patterns.

For most of the comparison methods, the proposed method performed much better.
For BOSS and CID, the proposed method only had a small advantage and for DTD$_C$, the proposed method had fewer wins.
This demonstrates that the proposed method is not only effective at providing an effective warping method, but also a robust distance measure for classification. 

\section{Experiments in Plug-in Scenario \label{sec:exp2}}
We further conducted the experimental evaluation of the proposed deep attentive time warping in the plug-in scenario of Fig.~\ref{fig:scenarios}~(b). The aims of this experiment are twofold. First, we evaluate the proposed method in a more practical task that needs representation learning in addition to time warping. Second, we compare the performance of the proposed method with state-of-the-art learnable time warping methods for the task.\par
We focused on online signature verification, which is a task to decide whether a test signature is a genuine signature or a skilled forgery (imitated by a forger). The reasons for using this task are as follows. The first and the most important reason is that several learnable time warping methods have been applied to the same public dataset, MYCT-100. As far as the authors know, there are no other common tasks to which various learnable time warping methods are applied. Second, this task requires representation learning; recent performance improvements on online signature verification owe to representation learning.
Third, this task still needs further improvement; even though recent methods achieve 1.0\% equal error rate (EER), verification error should be further minimized because of its expected reliability.   
\subsection{Online Signature Dataset \label{subsec:ex2_dataset}}

\begin{figure}[t]
\centering
\subfloat[Reference signature]{
\shortstack{
\includegraphics[width=.28\linewidth]{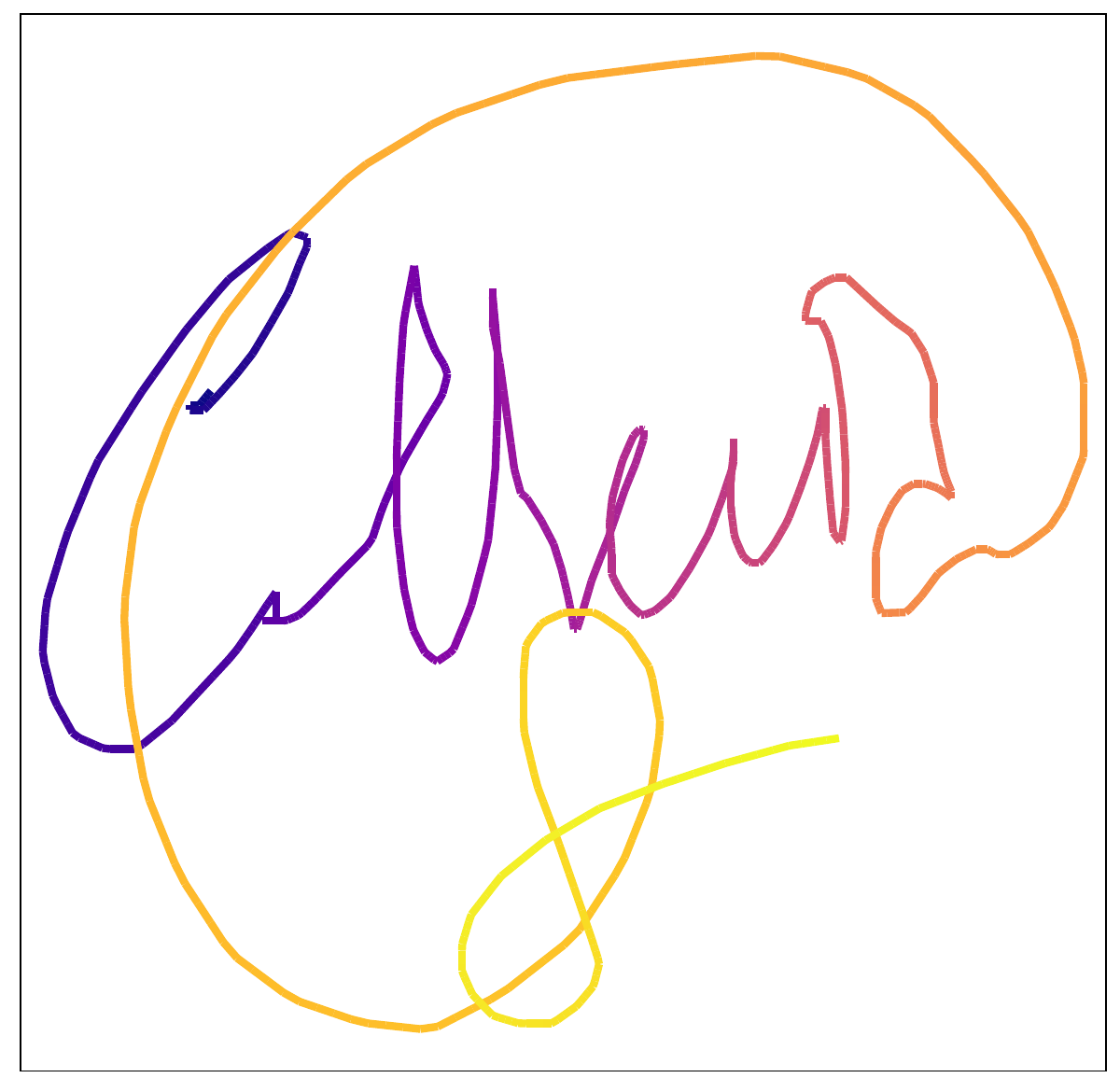}
\\[0ex]
\includegraphics[width=.3\linewidth]{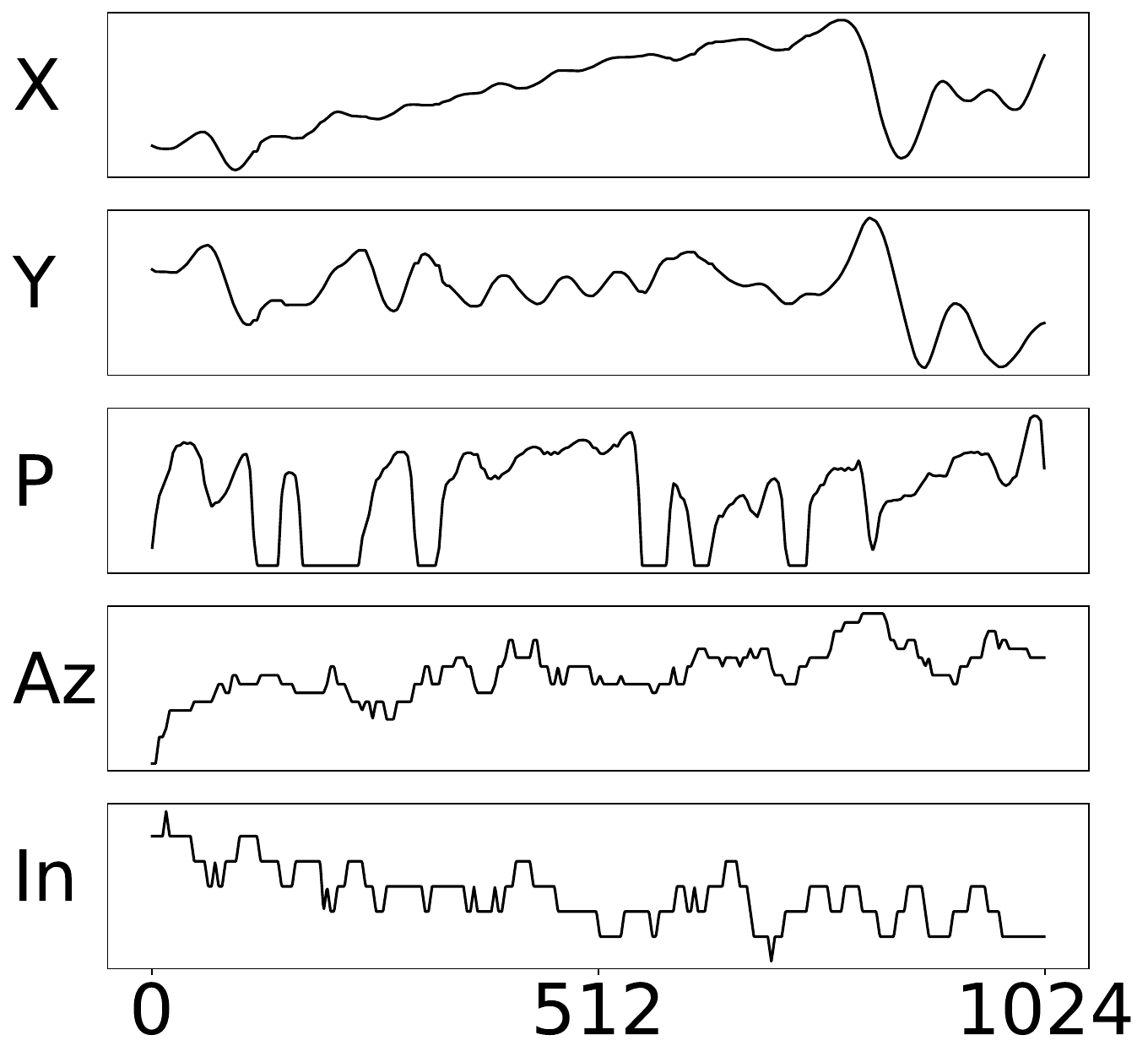}}
}
\hfill
\subfloat[Genuine signature]{
\shortstack{
\includegraphics[width=.28\linewidth]{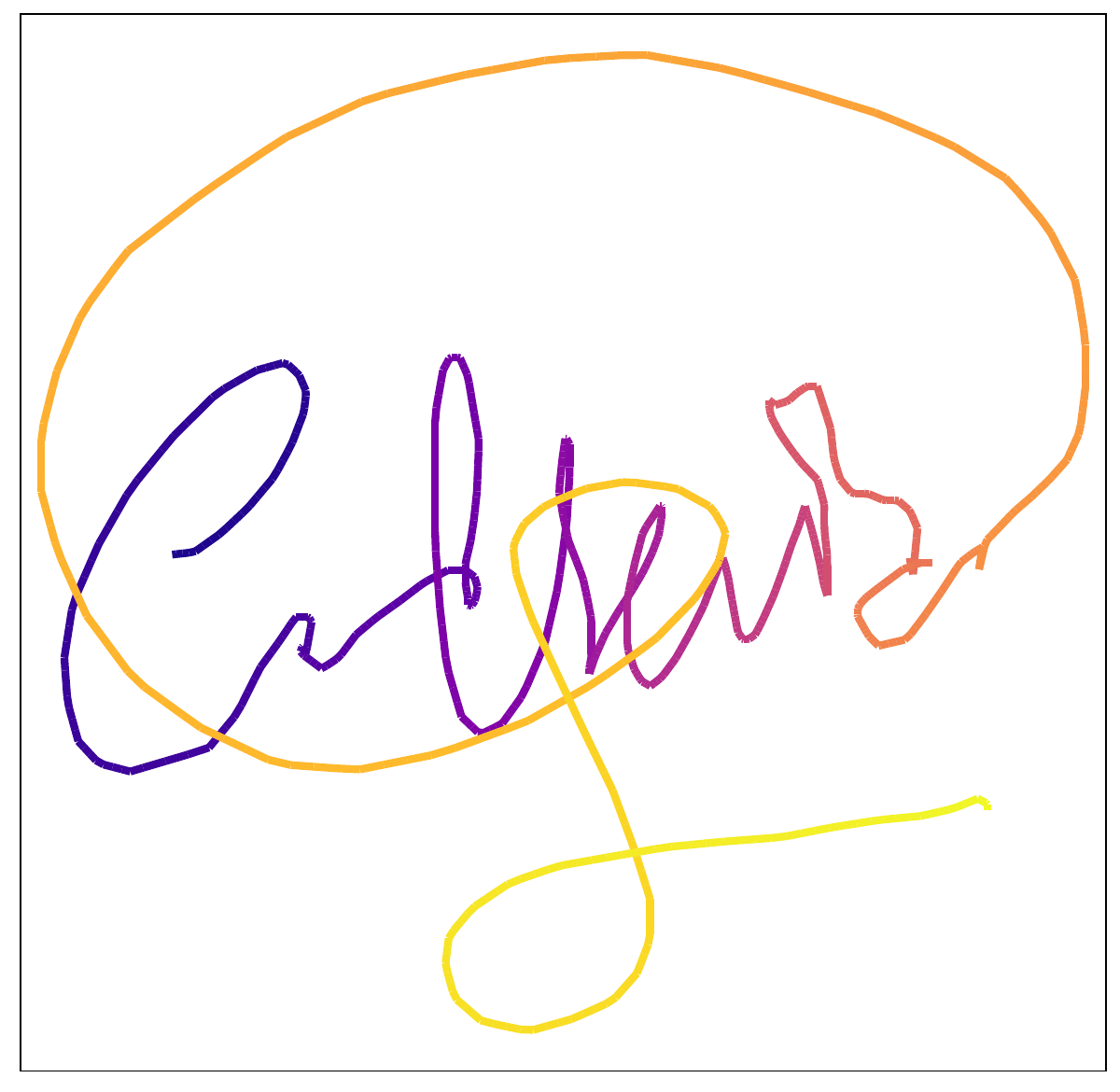}
\\[0ex]
\includegraphics[width=.3\linewidth]{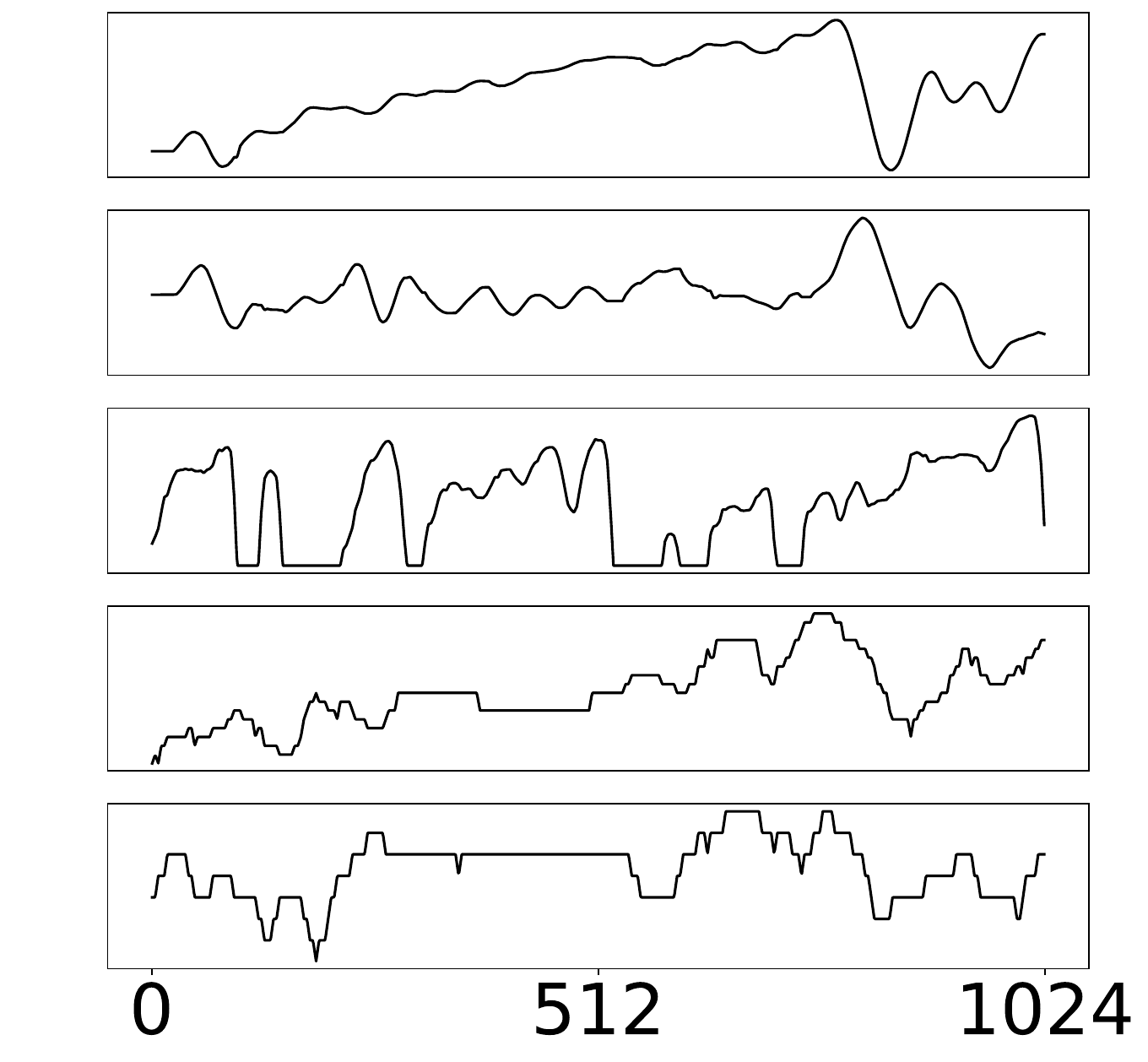}}
}
\hfill
\subfloat[Skilled forgery signature]{
\shortstack{
\includegraphics[width=.28\linewidth]{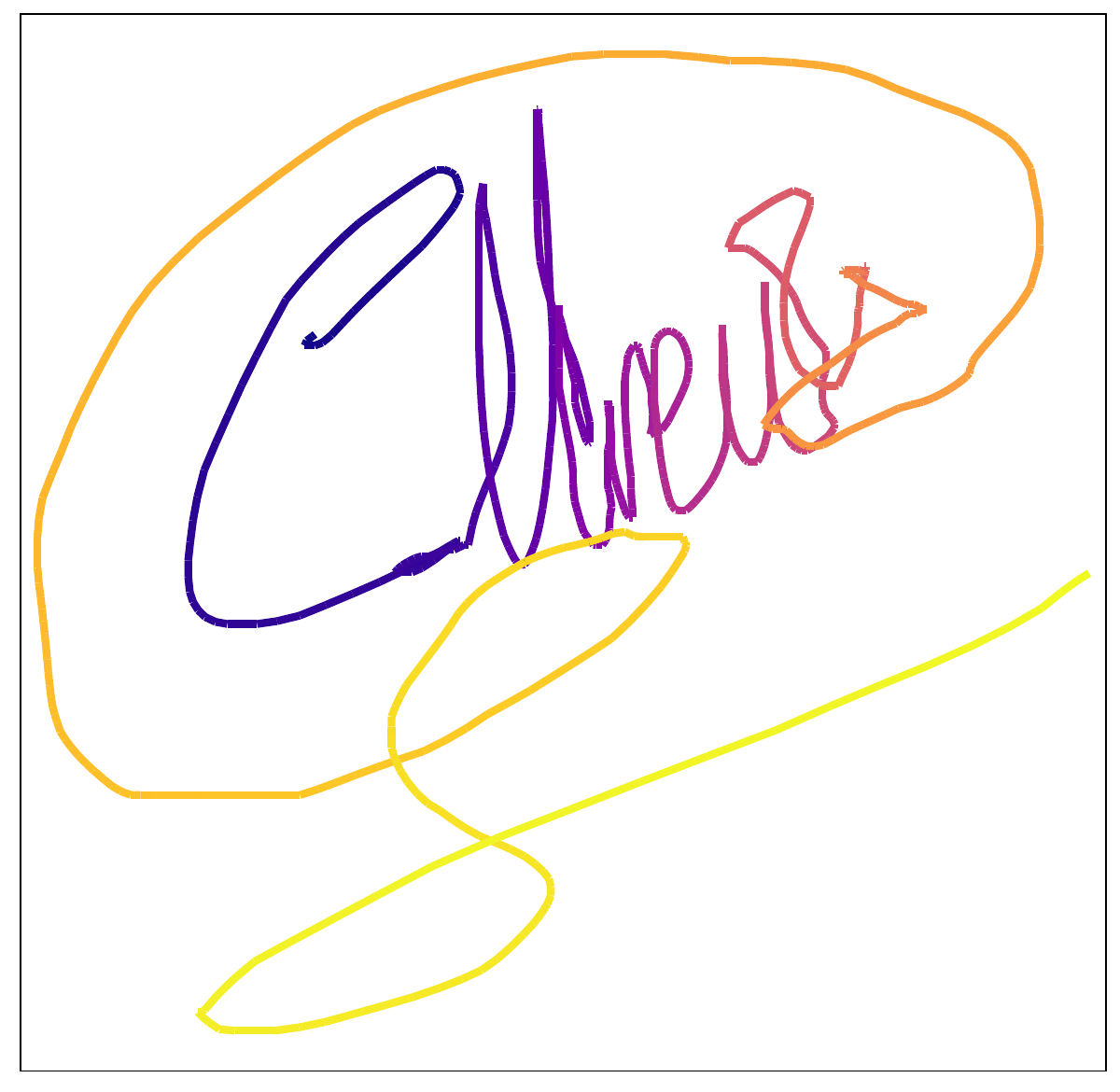}
\\[0ex]
\includegraphics[width=.3\linewidth]{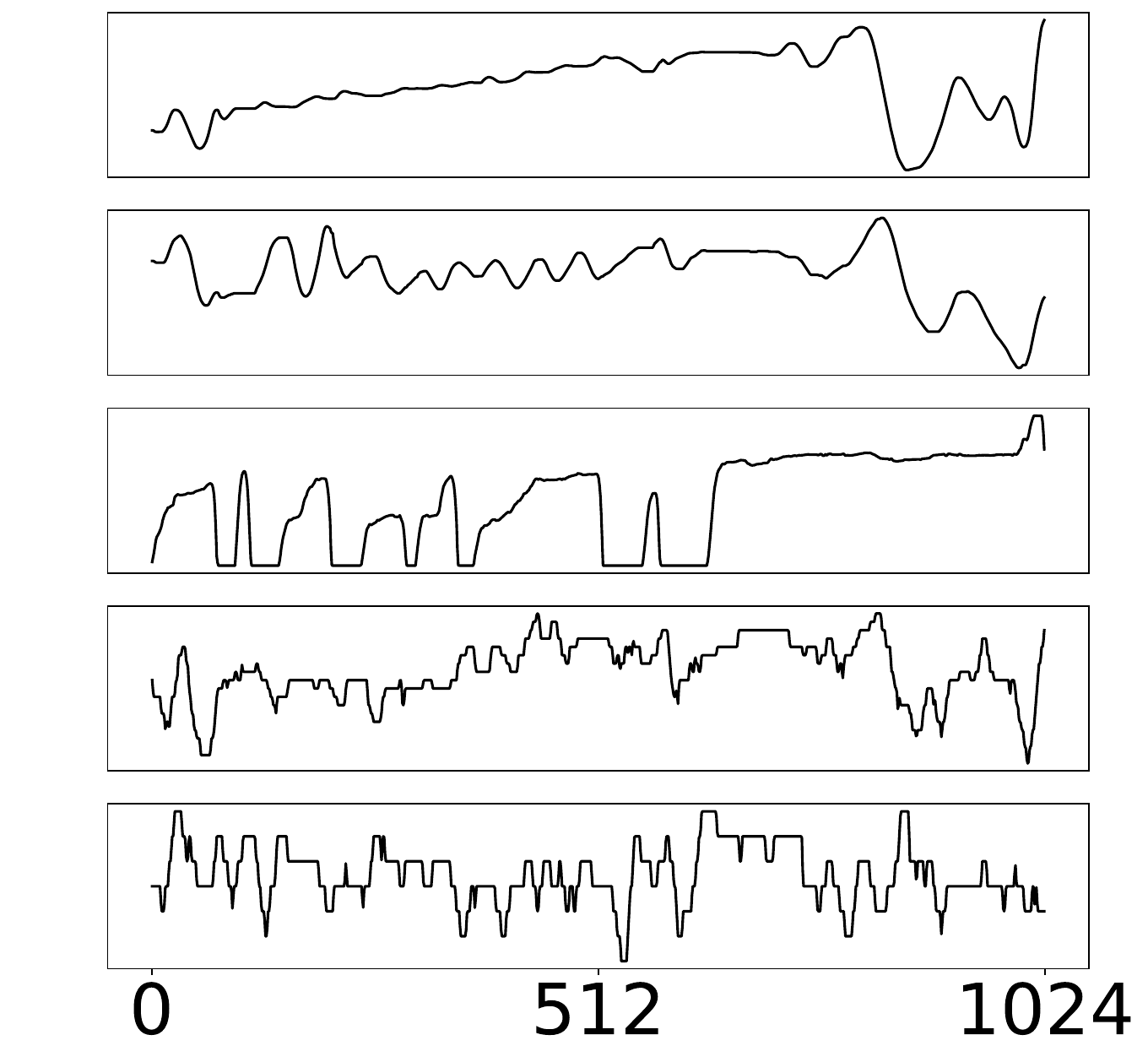}}
}
\caption{Examples of online signatures in MCYT-100.}
\label{fig:mcyt_100_ex}
\end{figure}

As one of the most common datasets for online signature verification, we use MCYT-100 dataset~\cite{MCYT03}. This dataset contains 100 subjects, each having 25 genuine signatures and 25 skilled forgeries. Fig.~\ref{fig:mcyt_100_ex} shows several examples from the dataset.  Each signature has five channels: 2D coordinates, pressure, azimuth, and altitude angles of the pen tip. We resized the temporal length to 1,024 by following the experimental setup of the comparative methods (such as PSN and DDTW).\par
According to the tradition of the task, we conduct multiple experiments under different train-test ratios. Specifically, the first $\eta\%$ of subjects ($\eta\in\{50, 60, 70, 80,\allowbreak 90\}$) were used for training and the remaining subjects for testing. For testing, the first five genuine signatures of each subject in the test set were used as reference signatures. The remaining genuine signatures and all the skilled forgeries were used as test signatures. For each test signature, its distances to the corresponding five reference signatures were averaged. Based on this averaged distance, all test signatures of all subjects were sorted to form a ranking list. EER was finally calculated as the traditional evaluation metric of the task.
\subsection{Comparative methods \label{subsec:ex2_comparative}}
In the plug-in scenario experiment, we used two elementary methods and three state-of-the-art methods.
The former methods are DTW~\cite{DTW78} and simple Siamese Network (Siamese).  DTW takes either raw signatures or handcrafted features~\cite{Martinez-DiazFK14} as input. Siamese was learned with either a global contrastive loss~\cite{HadsellCL06} or a local embedding loss~\cite{wu2019psn}. \par
\sloppy{
The state-of-the-arts methods are Prewarping Siamese Network (PSN)~\cite{wu2019psn}, Time-Aligned Recurrent Neural Networks (TARNN)~\cite{tarnn21}, and Deep DTW (DDTW)~\cite{wu2019ddtw}.}
In the latter method, DTW is embedded in a metric learning framework to realize learnable time warping. More specifically, the original PSN and TARNN use DTW before Siamese networks, whereas the original DDTW after. As the Siamese networks, PSN and DDTW employ CNN, whereas TARNN employs RNN.\par
For the comparison, we plugged the proposed method in the above methods and observed how the performance changed. Specifically, we replace the DTW module in PSN, DDTW, and TARNN by the proposed method and then train the entire network by the process of \ref{subsec:ex2_implement}.
\subsection{Implementation details\label{subsec:ex2_implement}}

\begin{figure}[t]
\centering
\includegraphics[width=1.0\linewidth]{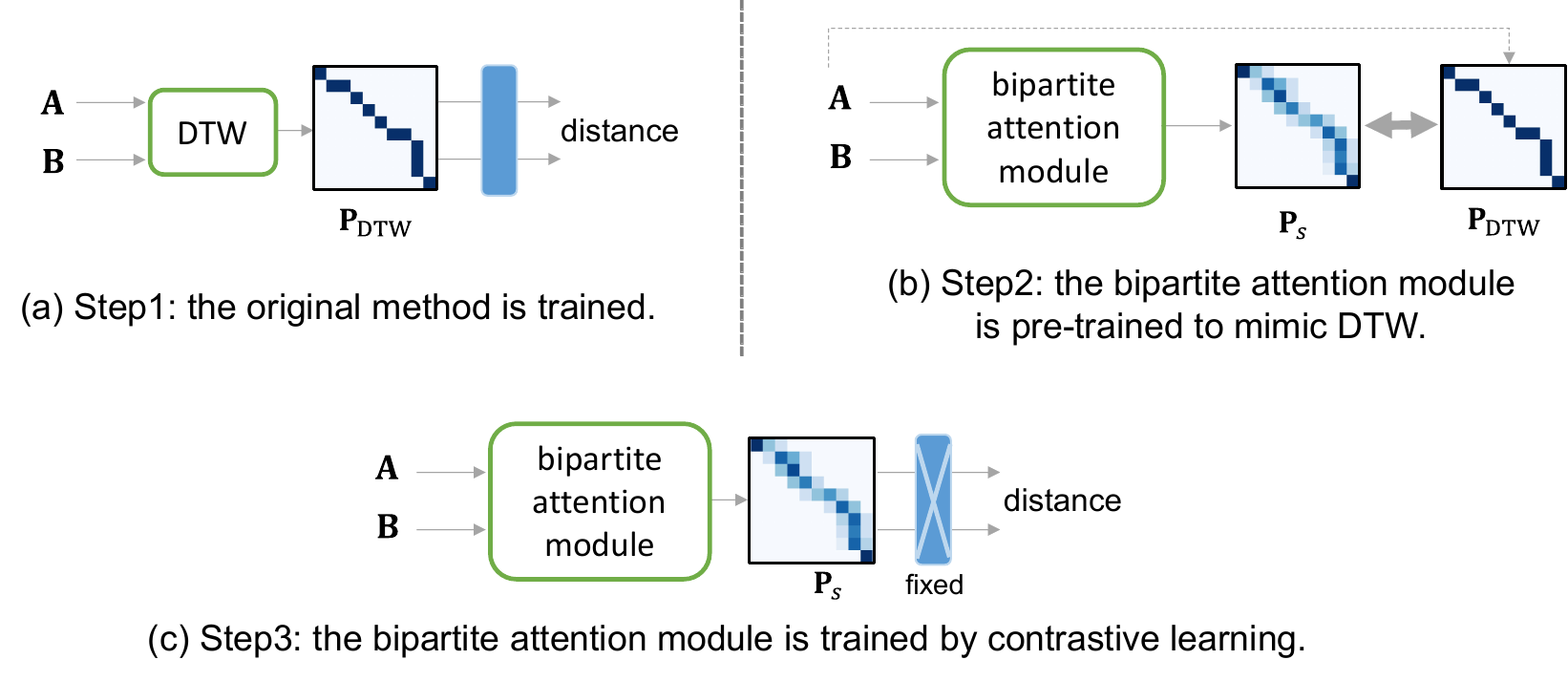}
\caption{Three-step training process for the plug-in scenario (for PSN/TARNN). Each blue box is the Siamese network for contrastive representation learning. Note that for DDTW, the Siamese network is placed {\em before} the bipartite attention module.}
\label{f:plug-in_training-step}
\end{figure}

Fig.~\ref{f:plug-in_training-step} shows
the three-step training process for the deep attentive time warping plugged in PSN, TARNN, and DDTW.
First, the original model with the standard DTW is trained with its original loss functions. (The details can be found in the original papers~\cite{wu2019psn,tarnn21,wu2019ddtw}.)
Second, the bipartite attention module is pre-trained with the loss (\ref{e:pre-training}), for providing a similar attention weight matrix to the DTW result. Finally, the bipartite attention module is trained with the entire network, while keeping the weights of the Siamese network. It is theoretically possible to train the entire network in an end-to-end manner. However, our preliminary trials proved that this three-step process gives more stable results.\par 

The network architecture of the bipartite attention module is the same as the stand-alone scenario. The learning rate was chosen from $0.1$, $0.01$, and $0.001$ in Steps 1 and 3, and set to $0.001$ in Step 2. The hyper-parameter $\tau$ in the contrastive loss was set to $1.4$ by using the validation set. Adam was used as the optimizer. The training was conducted up to $10,000$ iterations in Steps 1 and 3, and up to $1,000$ iterations in Step 2. The batch size was set to $30$, where $10$ are the same-class pairs and $20$ are the different-class pairs).

\subsection{Results\label{subsec:ex2_result}}

\begin{table}[t]
\caption{EERs (\%) of online signature verification on MCYT-100. EER in \textcolor{red}{red} and \textcolor{blue}{blue} indicate the least and the second least rates, respectively.}
\label{tab:verification_result}
\centering
\begin{tabular*}{\linewidth}{@{\extracolsep{\fill}}lccccc}
\toprule
& \multicolumn{5}{c}{Percentage ($\eta$\%) of Training Data} \\
\cmidrule(lr){2-6}
~Method & 90 & 80 & 70 & 60 & 50 \\
\midrule
~DTW~\cite{DTW78,Martinez-DiazFK14} & 4.00 & 3.00 & 4.17 & 4.37 & 4.60 \\
~~~w/ raw signatures & 5.00 & 6.25 & 5.73 & 6.37 & 6.96 \\
~Siamese & 5.50 & 6.80 & 6.27 & 7.33 & 8.40 \\
~~~w/ local embedding loss~\cite{wu2019psn} & 3.50 & 3.40 & 3.75 & 3.75 & 5.50 \\
\midrule
~PSN~\cite{wu2019psn} &  1.50 & 2.25 & 3.17 & 2.75 & 3.00 \\ 
~~~+ ours (plug-in) & 1.00 & \textcolor{red}{1.75} & \textcolor{red}{2.33} & \textcolor{red}{2.13} & 2.70 \\ 
~~~~~~~w/o pre-training & 1.00 & 2.50 & 3.67 & 3.50 & 4.10 \\ 
\midrule
~TARNN~\cite{tarnn21} & 1.00 & 3.00 & 3.50 & 4.25 & 4.50 \\ 
~~~+ ours (plug-in) & \textcolor{red}{0.50} & 2.25 & 2.67 & 2.88 & 2.80 \\
~~~~~~~w/o pre-training & 1.50 & 2.50 & 3.17 & 3.25 & 5.00 \\ 
\midrule
~DDTW~\cite{wu2019ddtw} & 1.00 & 2.20 & 2.53 & 2.25 & \textcolor{blue}{2.40}  \\
~~~+ ours (plug-in) & \textcolor{red}{0.50} & \textcolor{blue}{2.00} & \textcolor{red}{2.33} & \textcolor{red}{2.13} & \textcolor{red}{2.20} \\ 
~~~~~~~w/o pre-training & 1.50 & 4.00 & 4.83 & 3.50 & 3.90 \\ 
\bottomrule
\end{tabular*}
\vspace{2mm}
\end{table}

Table~\ref{tab:verification_result} shows the EERs on MCYT-100. The accuracy of the state-of-the-art methods, i.e., PSN, TARNN, and DDTW,  has all been improved by replacing DTW with the proposed method.
This proves that the proposed method is consistently effective as a plug-in to existing learnable time warping frameworks. In addition, from the comparison between PSN and DDTW, the plug-in location (i.e., before or after the representation learning module) is not very important. \par
Removing pre-training from the proposed method degrades the performance significantly. This fact confirms the necessity of the proposed pre-training method in improving discriminative power and stabilizing the inference of bipartite attention matrices.\par
\begin{figure}[t]
\centering
\subfloat[DDTW]{
\includegraphics[width=.18\linewidth]{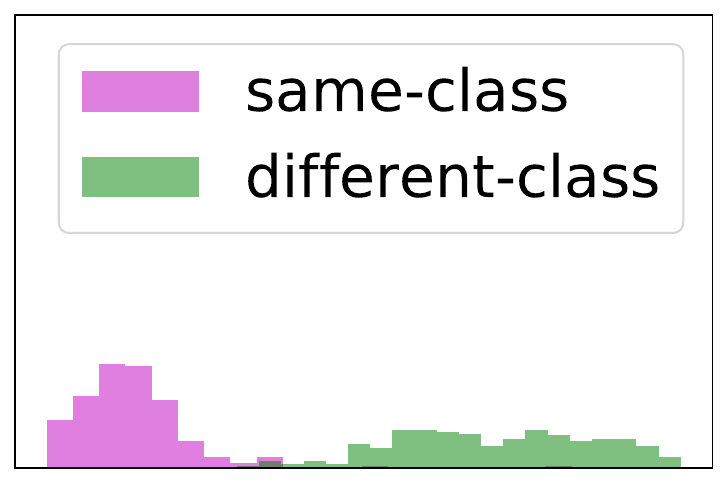}
\includegraphics[width=.18\linewidth]{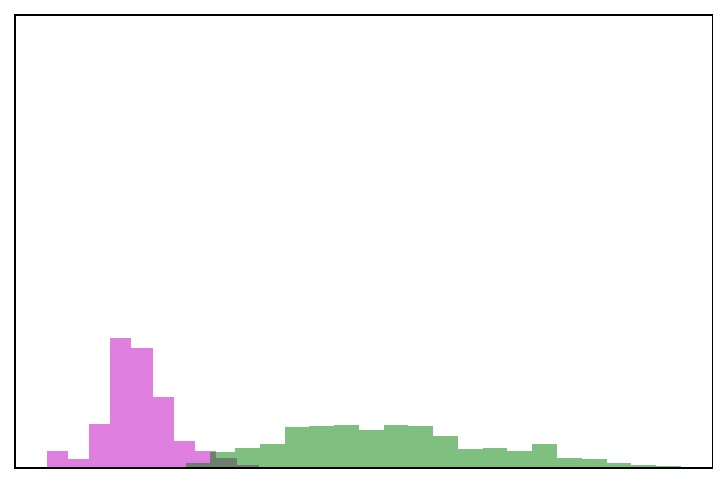}
\includegraphics[width=.18\linewidth]{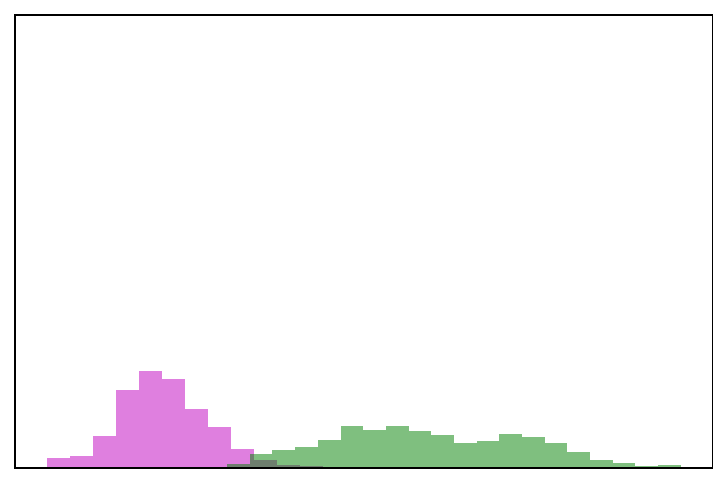}
\includegraphics[width=.18\linewidth]{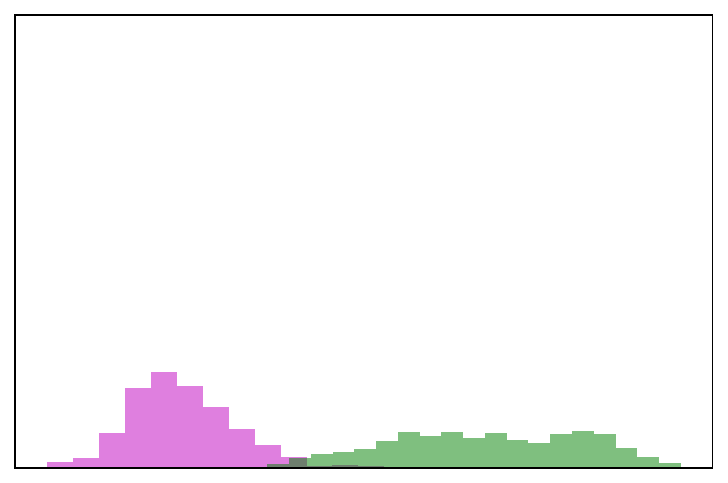}
\includegraphics[width=.18\linewidth]{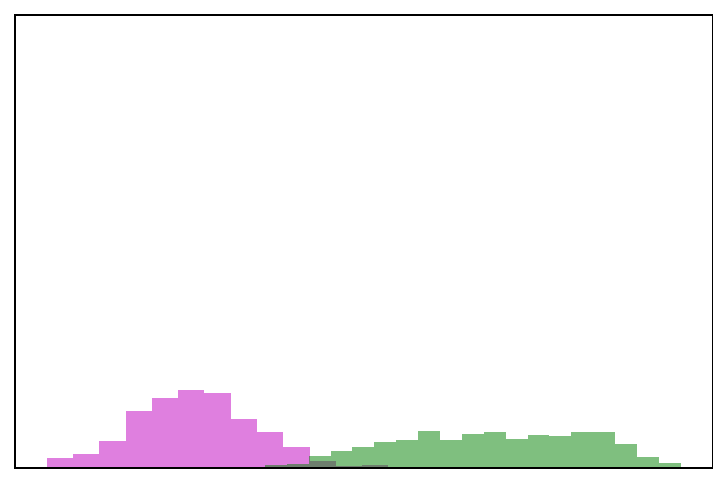}
}\\
\centering
\subfloat[ours]{
\includegraphics[width=.18\linewidth]{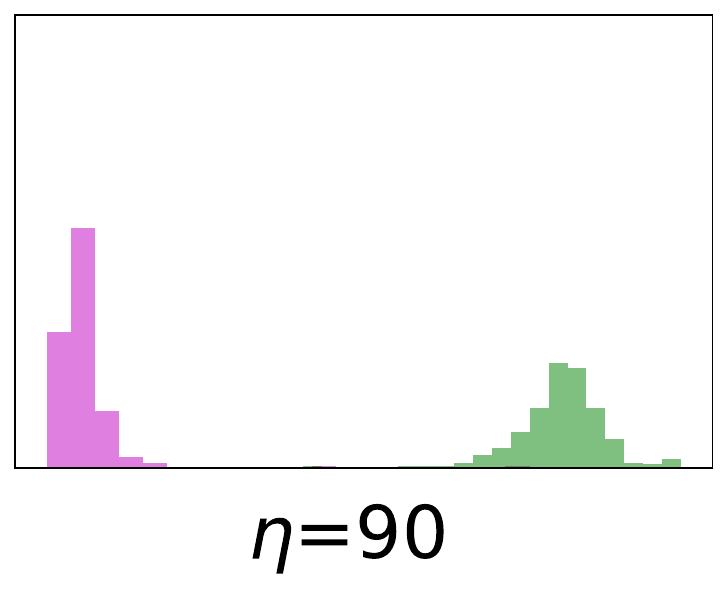}
\includegraphics[width=.18\linewidth]{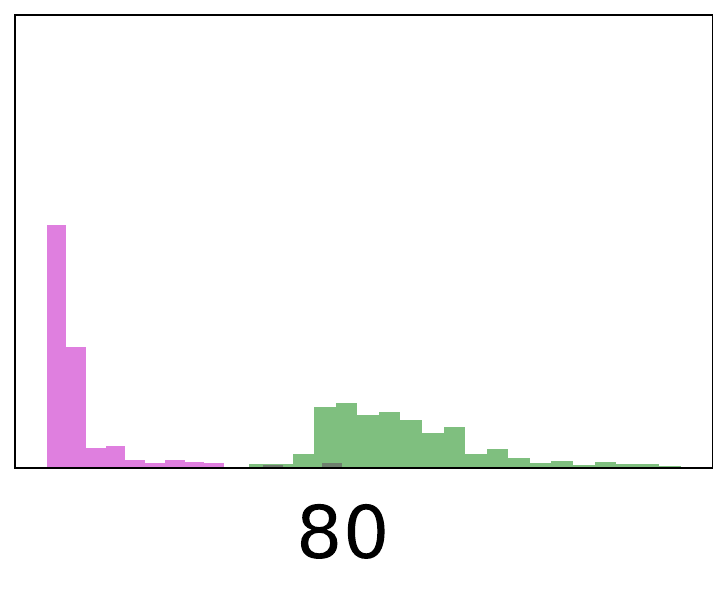}
\includegraphics[width=.18\linewidth]{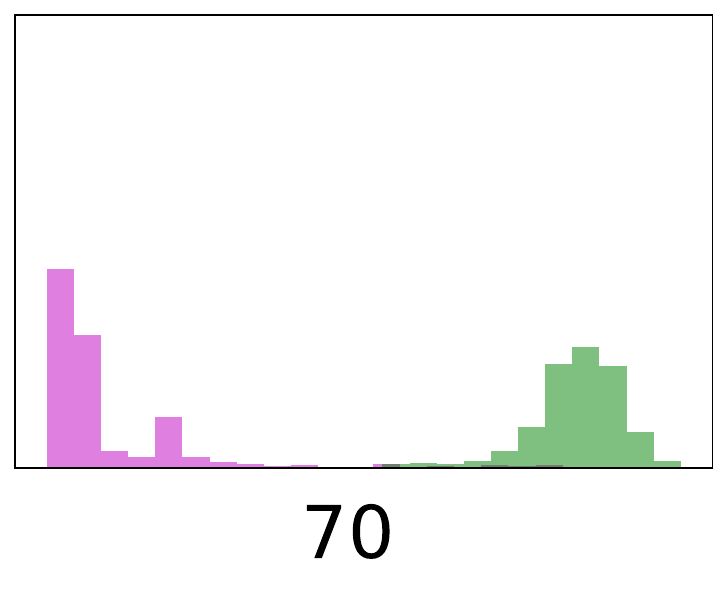}
\includegraphics[width=.18\linewidth]{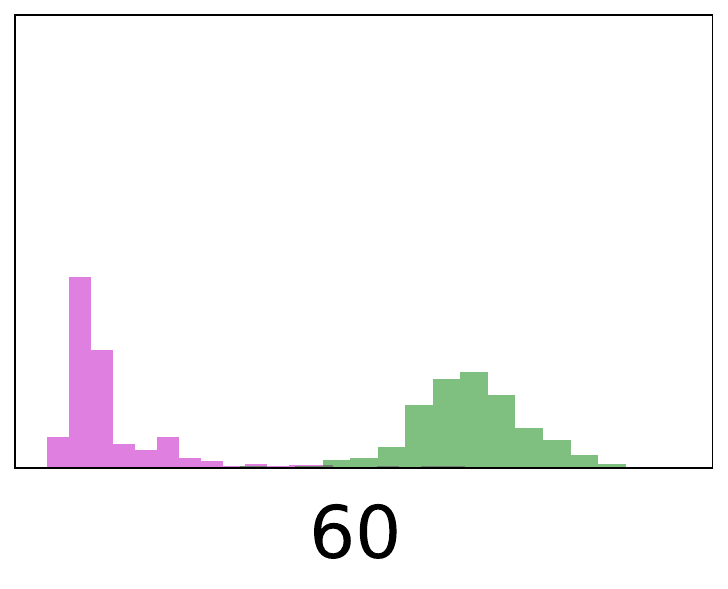}
\includegraphics[width=.18\linewidth]{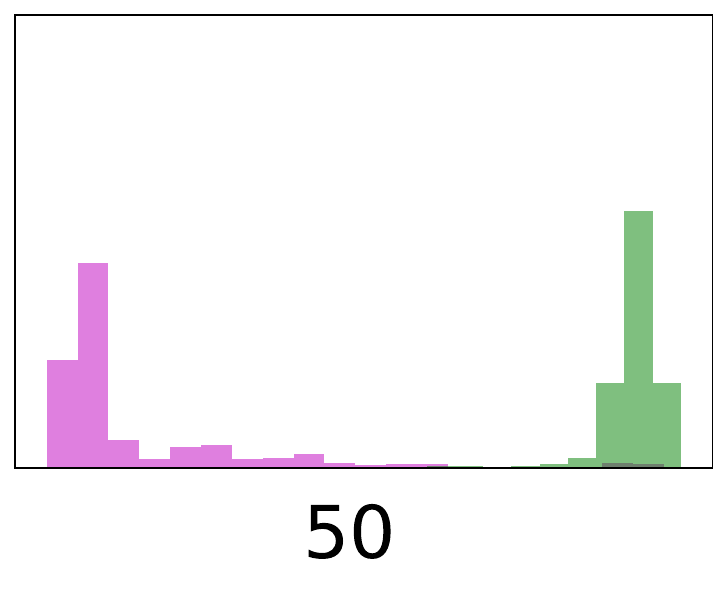}
}
\caption{Distance histograms by the proposed method and DDTW~\cite{wu2019ddtw} on MCYT-100. The horizontal axis is the distance and the vertical axis is normalized.}
\label{fig:hist_dtw_signature}
\end{figure}
Fig.~\ref{fig:hist_dtw_signature} shows the distance histograms of the same-class and different-class pairs
by the proposed method and DDTW. The proposed method shows a much smaller overlap than DDTW. DDTW focuses only on contrastive representation learning; in contrast, the proposed method trains the attention weight matrix (i.e., soft-correspondence) in contrastive learning, in addition to representation learning. This makes the proposed method more discriminative than DDTW.\par

\section{Conclusion\label{sec:conclusion}}
In this paper, we proposed a novel neural network-based time warping method, called deep attentive time warping. The proposed method is based on a new attention module, called the bipartite attention module, between two time series inputs. The module is trained by contrastive metric learning to achieve a learnable and task-adaptive time warping and to improve the trade-off between robustness against time distortion and discriminative power. 
The effectiveness of the proposed method was confirmed through two scenarios. 
The first was a stand-alone scenario, where the proposed method was used as a learnable time warping method and compared with the standard DTW and other time warping methods. Through qualitative and quantitative evaluations with Unipen and UCR datasets, the expected effectiveness was confirmed. The second was a plug-in scenario, where the proposed method is embedded in neural network-based metric learning frameworks with representation learning. Through a comparative study with state-of-the-art learnable time warping methods, the effectiveness of the proposed method was further confirmed.\par
The limitations of this paper are as follows. First, in the current framework, the regulation of the warping flexibility relies on the pre-training to mimic the standard DTW and the soft constraints implicitly imposed by the trained bipartite attention module; this means no explicit penalty for the violation of several reasonable regulations, such as monotonicity and continuity of warping. Although we confirmed the performance superiority over the standard DTW with those warping regulations, there is still a possibility that the introduction of some explicit regulations will further improve the performance. 
 Second, we optimize the network architectures according to the characteristics of the dataset. In this paper, we used the same architecture for all UCR2015 datasets, and therefore the performance degrades for too long or too short time-series samples, as revealed by the analysis in Section~\ref{sec:quantitative-UCR}. From a practical viewpoint, architecture optimization for better performance will be an important future work.
Third, 

we have not directly utilized the soft-correspondence (represented by the attention weight matrix) in the final distance evaluation. In fact, the attention weight matrix can be seen as a novel feature showing the relationship between two time series, and therefore we can extract some useful features from it for final distance evaluation and/or final decision making.

\section*{Acknowledgments}
This work was partially supported by MEXT-Japan (Grant No. J17H06100 and J22H00540).

\bibliography{bibfile}

\clearpage

\begin{appendices}

\section{Result on datasets from UCR2018 \label{app:UCR2018}}
In addition to results of UCR2015 in Table~\ref{tab:UCR}, we conducted the same experimental evaluation for UCR2018~\cite{UCRArchive2018}. In UCR2018, we found six datasets that satisfy the same conditions (the sample size, the time length, and the fixed-length samples) as the UCR2015 experiment.
Table~\ref{tab:ucr2018} shows the error rate of the datasets. Except for ``Crop,'' which contains many training samples beneficial for 1NN classification by DTW, ours could achieve the best accuracy (even perfect).

\begin{table}[t]
\centering
\caption{Error rate (\%) on additional six datasets from UCR2018. Error rates in \textcolor{red}{red} and \textcolor{blue}{blue} indicate the least and the second least rates, respectively.}
\label{tab:ucr2018}
\scalebox{0.6}[0.53]{
\begin{tabular}{lrrrrrrrrr}
\toprule
Dataset              & Train & Test  & Class & Length & DTW   & w-DTW & s-DTW & ours  & w/o pre-train \\
\midrule
Crop                     & 7200  & 16800 & 24    & 46     & 33.00 & \textcolor{red}{28.83} & \textcolor{blue}{31.46}  & 37.13 & 32.13         \\
FreezerRegularTrain      & 150   & 2850  & 2     & 301    & 10.00 & 9.30  & 6.84  & \textcolor{red}{0.00}  & \textcolor{red}{0.00}          \\
GunPointAgeSpan          & 135   & 316   & 2     & 150    & 8.00  & 3.48  & 1.58  & \textcolor{red}{0.00}  & \textcolor{red}{0.00}          \\
GunPointMaleVersusFemale & 135   & 316   & 2     & 150    & \textcolor{red}{0.00}  & 2.53  & 1.58  & \textcolor{red}{0.00}  & \textcolor{red}{0.00}          \\
GunPointOldVersusYoung   & 136   & 315   & 2     & 150    & 16.00 & 3.49  & \textcolor{red}{0.00}  & \textcolor{red}{0.00}  & \textcolor{red}{0.00}          \\
PowerCons                & 180   & 180   & 2     & 144    & 12.00 & 7.78  & 9.44  & \textcolor{red}{0.00}  & \textcolor{red}{0.00}         \\
\bottomrule
\end{tabular}
}
\end{table}

\end{appendices}


\end{document}